%% file: main.tex
\documentclass[10pt,twocolumn,letterpaper]{article}

\usepackage{cvpr}
\usepackage{times}
\usepackage{epsfig}
\usepackage{graphicx}
\usepackage{amsmath}
\usepackage{amssymb}
\usepackage{algorithm}
\usepackage{algorithmic}
\usepackage{soul}
\usepackage{caption, subcaption}

\usepackage[breaklinks=true,bookmarks=false]{hyperref}

\cvprfinalcopy 

\def\cvprPaperID{2485} 

\input{math_defs.tex}

\begin{document}

\title{A Graduated Filter Method for Large Scale Robust Estimation}

%
%

\author{Huu Le and Christopher Zach\\
Chalmers University of Technology, Sweden\\
{\tt\small \{huul, zach\}@chalmers.se}
}

\maketitle

\begin{abstract}
Due to the highly non-convex nature of large-scale robust parameter estimation, avoiding poor local minima is challenging in real-world applications where input data is contaminated by a large or unknown fraction of outliers.  
In this paper, we introduce a novel solver for robust estimation that possesses a strong ability to escape poor local minima. Our algorithm is built upon the class of traditional graduated optimization techniques, which are considered state-of-the-art local methods to solve problems having many poor minima.
The novelty of our work lies in the introduction of an adaptive kernel (or residual) scaling scheme, which allows us to achieve faster convergence rates.
Like other existing methods that aim to return good local minima for robust estimation tasks, our method relaxes the original robust problem, but adapts a filter framework from non-linear constrained optimization to automatically choose the level of relaxation.
Experimental results on real large-scale datasets such as bundle adjustment instances demonstrate that our proposed method achieves competitive results. \footnote{Our C++ implementation is available at~\url{https://github.com/intellhave/ASKER}.}

\end{abstract}

\input{tex/1-introduction.tex}

\input{tex/2-related.tex}
\input{tex/3-formulation.tex}

\input{tex/4-approach.tex}
\input{tex/5-results.tex}

\section{Conclusion and Future Work}

In this work we propose a method for large-scale robust estimation that uses an optimization-driven schedule to steer the difficulty of intermediate optimization tasks.
This puts out method in contrast to graduated optimization techniques such as graduated non-convexity, which always uses a monotone schedule from easy to successively harder problem instances.
By using an adaptive schedule, in our experiments the proposed method achieves a good balance between fast decrease of the target objective and reaching a competitive local minimum.

Since filter methods are one framework to relax (or to smooth) difficult optimization problems in a well-justified way, future work will investigate into further applications of this technique. Another direction for future work is to further leverage the problem structure, such as the bi-partite nature of unknowns in bundle adjustment.

\section*{Acknowledgment}
\noindent This work was partially supported by the Wallenberg AI, Autonomous
Systems and Software Program (WASP) funded by the Knut and Alice
Wallenberg Foundation.

{\small
\bibliographystyle{ieee_fullname}
\bibliography{main}
}

\section*{Supplementary Material}
\input{supp/more_results.tex}
\input{supp/construction_visual}
\input{supp/step_computation.tex}
\input{supp/param_choice.tex}
\input{supp/convergence.tex}
\input{supp/gradient.tex}

\end{document}

%% file: math_defs.tex
\newcommand{\cF}{\mathcal{F}}

\newcommand{\br}{\mathbf{r}}
\newcommand{\bx}{\mathbf{x}}

\newcommand{\bX}{\mathbf{X}}

\newcommand{\bs}{\mathbf{s}}
\newcommand{\bp}{\mathbf{p}}

\newcommand{\bR}{\mathbf{R}}
\newcommand{\bt}{\mathbf{t}}
\newcommand{\bH}{\mathbf{H}}

\newcommand{\bu}{\mathbf{u}}

\newcommand{\bg}{\mathbf{g}}

\newcommand{\bF}{\mathbf{F}}
\newcommand{\bI}{\mathbf{I}}
\newcommand{\tdf}{\tilde{f}}
\newcommand{\tdh}{\tilde{h}}

\newcommand{\bbR}{\mathbb{R}}

\newcommand{\bbF}{\mathbb{F}}

\newcommand{\bhr}{\hat{\mathbf{r}}}
\newcommand{\bJ}{\mathbf{J}}

\newcommand{\kernel}{\psi}

\newcommand{\btheta}{\boldsymbol{\theta}}

%% file: tex/1-introduction.tex
\section{Introduction}
Robust parameter estimation plays a crucial role in many computer vision tasks, ranging from low-dimensional model fitting (e.g., fundamental or essential matrix estimation~\cite{hartley2003multiple}) to large-scale instances in very high dimensional space, which may contain hundreds of thousands of measurements (e.g., pose graph optimization~\cite{kummerle2011g}, SLAM~\cite{mur2015orb} and bundle adjustment~\cite{snavely2006photo}). 
When the input data is relatively clean with a low percentage of gross outliers, the set of optimal parameters can be easily obtained under the maximum likelihood framework, i.e., by minimizing the sum of the squared residuals~\cite{triggs1999bundle}, and the cost can be optimized using popular off-the-shelf non-linear least squares solvers (e.g., Ceres~\cite{ceres-solver}). 
However, under the presence of a large fraction of gross outliers, standard least-squares fit often yields results that are biased toward outlying measurements. To achieve robustness in low-dimensional settings, randomized approaches such as RANdom SAmple Consensus (RANSAC)~\cite{fischler1981random} and its variants~\cite{chum2003locally,chum2005matching} are preferred. 
Deterministic algorithms that provide global~\cite{chin15} or local solutions~\cite{le2017exact, cai2018deterministic} to improve RANSAC also exist. 
However, they are inapplicable in large-scale settings, where a class of M-estimators~\cite{huber81} must be employed.
Under this framework, if the fraction of outliers is small, then convex $\ell_1$ or Huber kernels can be sufficient. 
However, in order to achieve maximum robustness for the parameter estimation task with a large outlier ratio, a quasi-convex kernel is usually employed, which in most cases leads to non-convex problems with many sub-optimal local minima and flat regions in parameter space~\cite{zach2014robust}. 
Solving such non-convex problems is well-known to be challenging as it is very likely for an algorithm to converge to a poor local minimum. Our work proposes a new algorithm to address this problem.

A number of optimization schemes have been proposed to tackle the high non-convexity of robust estimation, and~\cite{zach2018descending} evaluates some of the promising methods. 
Among these, graduated optimization, which is often referred as graduated non-convexity (GNC) in the computer vision community, shows to be the most promising approach due to its competitive ability to escape poor solutions. Therefore, they have been used widely in many robust fitting applications (e.g.~\cite{blake1987visual,mobahi2015link,zach2018descending,yang2019graduated}). 
However, the use of GNC requires a careful design of the graduated optimization schedule, which requires prior knowledge about the problem.
A wrong schedule may cause either unnecessarily long run time in several easy problem instances, where basic techniques that provide fast convergence such as Iteratively Re-weighted Least Squares (IRLS) are sufficient, or undesirable results as local minima are not effectively avoided (as demonstrated in Figure~\ref{fig:gnc_limitation}).

\paragraph{Contributions} To address the above problems, we introduce in this paper a new algorithm for large-scale robust estimation that is as competitive as GNC, but does not require a fixed optimization schedule. To achieve such goal, we propose to consider the scale parameters as variables that are jointly optimized with the original parameters. The introduction of the scale parameters to the original problem results in a constrained optimization problem, which can then be efficiently solved using a novel adaptation of the filter method~\cite{fletcher2002nonlinear}. Experimental results on several large-scale bundle adjustment datasets demonstrate that our method provides competitive objective values as well as convergence rates compared to existing state-of-the-art methods for robust estimation.


%% file: tex/2-related.tex
\section{Related Work}

Iteratively Re-weighted Least Squares (IRLS~\cite{green1984iteratively}) is arguably the most popular method being used to optimize high-dimensional robust cost functions. The main idea behind this approach is to associate each measurement with a weight computed based on its current residual, then minimize an instance of weighted least squares. The weights are updated after each iteration and the process repeats until convergence. It has been demonstrated that with a proper initialization of weights, IRLS may provide competitive results~\cite{zach2019pareto}. However, for more complex problems, the returned solutions are usually not satisfactory as it is very easy for IRLS to be trapped in a poor local minimum. 

To address the non-convexity of robust estimation, Zach~\cite{zach2014robust} leveraged the half-quadratic minimization principle~\cite{geman1992constrained} and proposed to solve the problem in a ``lifted" domain, where the non-convex robust kernel is re-parameterized by a new function in a higher dimensional space. The reformulated robust estimation problem incorporates both the original parameters and newly introduced unknowns representing the confident weights of the measurements. By employing such lifting approach, the flat region in the robust kernels can be avoided by indirectly representing the robustness into the new lifted objective, which is less sensitive to poor local minima. Using the lifting mechanism, different formulations and schemes have also been introduced. In contrast to the Multiplicative Half-Quadratic (M-HQ) lifting approach proposed in~\cite{zach2014robust}, Additive Half-Quadratic (A-HQ) has also been introduced~\cite{geman1995nonlinear, zach2018multiplicative}. A double lifting method that combines M-HQ and A-HQ is also discussed in~\cite{zach2018multiplicative}. However, the above lifting approaches have some limitations. 
In particular,~\cite{zach2019pareto} demonstrates that the success of half-quadratic minimization relies on suitable initialization of confidence weights, and that M-HQ fails on problems with multiple ``competing'' residuals.


Besides lifting, another popular approach to tackle problems containing many poor local minima is to ``smooth" the objective using homotopy or graduation techniques~\cite{rose1998deterministic,dunlavy2005homotopy,mobahi2015link} such as Graduated Non-convexity (GNC~\cite{blake1987visual}). The underlying concept of graduated optimization is to successively approximate the original non-convex cost function by surrogate functions that are easier to minimize (i.e., leading to fewer local minima). In robust cost optimization, the surrogate functions may be chosen as a scaled version of the original robust kernel (see Sec.~\ref{sec:GNC}), which induces fewer local minima than the original cost. Graduated optimization and GNC have demonstrated their utility in several large-scale robust estimation problems by guiding the optimization process to relatively good local minima compared to other approaches such as IRLS or lifting variants~\cite{zach2018descending}.

%% file: tex/3-formulation.tex
\section{Background}
\subsection{Problem Formulation}
In this work, we are interested in large-scale robust estimation under the framework of M-estimators. Assume that we are given a set of $N$ measurements, and let us denote the residual vector induced by the $i$-th observation by $\br_i(\btheta) \in \bbR^p$, where the vector $\btheta \in \bbR^d$ contains the desired parameters. In robust cost optimization, we wish to obtain the optimal parameters $\btheta^*$ that solve the following program 
\begin{align}
    \btheta^* = \arg\min_{\btheta} \Psi(\btheta) & & \Psi(\btheta) :=  \sum_{i=1}^N\kernel (||\br_i(\btheta)||),
    \label{eq:robust_mean}
\end{align}
where $\kernel: \bbR \mapsto \bbR$ is a symmetric robust kernel that satisfies the following properties~\cite{geman1992constrained,zach2018descending}: $\kernel(0) = 0$, $\kernel''(0) = 1$, and the mapping $\phi:\bbR_0^+  \mapsto \bbR_0^+$ where $\phi(x) = \kernel(\sqrt{2z})$ is concave and monotonically increasing.
The problem~\eqref{eq:robust_mean} serves as a generic framework for several robust fitting tasks, in which the definitions of the parameters $\btheta$ and the residual vectors $\{\br_i(\btheta)\}$ depend on the specific application. For example, in robust metric bundle adjustment, the parameter vector $\btheta$ consists of the set of camera matrices $\{\bR_j, \bt_j\}_{j=1}^{N_v}$ 
together with the set of 3-dimensional (3D) points $\{\bX_k\}_{k=1}^{N_p}$ ($N_v$ and $N_p$ are the number of cameras and the number of points, respectively), and each residual vector $\br_{ij} \in \bbR^2$ is defined as
\begin{equation}
    \br_{ij}(\btheta) = \bu_{ij} - \pi(\bR_i\bX_j + \bt_i),
    \label{eq:bundle_residual}
\end{equation}
where $\pi:\bbR^3 \mapsto \bbR^2$ is defined as $\pi(\bX) = (X_1/X_3, X_2/X_3)$, and $\bu_{ij}$ is the 2D keypoint corresponding to the $j$-th 3D point extracted in image $i$.

The robust kernel $\kernel$ can be chosen from a wide range of functions (See~\cite{zach2018descending}). This choice usually affects the robustness and the convexity of the resulting optimization problem. For example, if $\kernel(x)$ is chosen such that $\kernel(x) = \frac{x^2}{2}$, one obtains the non-robust least squares estimate, which is easy to optimize but very sensitive to outliers. In this work, if not otherwise stated, we chose $\kernel$ to be the smooth truncated kernel,
\begin{equation}
    \kernel(r) = \left.
    \begin{cases}
        \frac{1}{2} r^2 \left(1- \frac{r^2}{2\tau^2}\right) & \text{if} \; r^2 \le \tau^2,\\
        \tau^2/4 & \text{otherwise.}
    \end{cases}\right.
\end{equation}

\subsection{Graduated Optimization and Its Limitations}
\label{sec:GNC}
\begin{figure}[ht]
    \centering
    \includegraphics[width = 0.85\columnwidth]{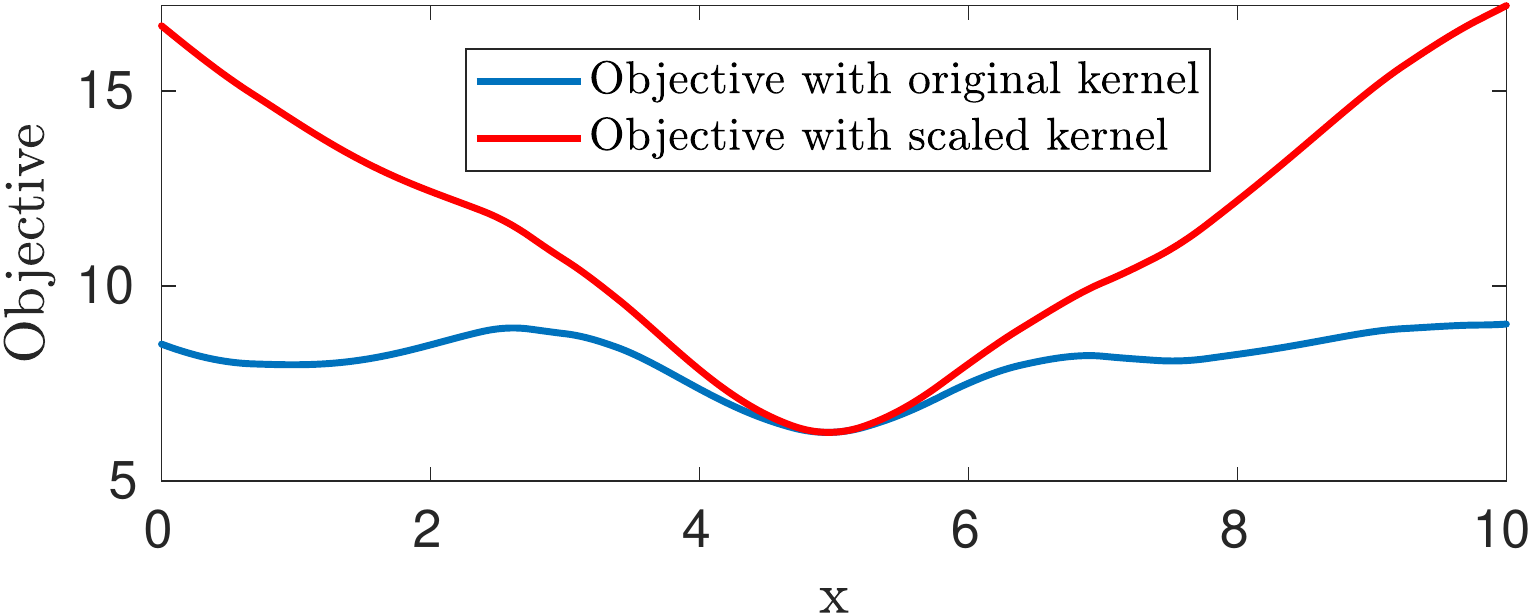}
    \caption{Illustration of a 1-d robust mean fitting problem, where the surrogate objective with scaled kernel (red) contains fewer local minima than the original cost (blue).}
    \label{fig:robust_mean}
\end{figure}

\begin{figure}[ht]
    \centering
    \includegraphics[width = 0.85\columnwidth]{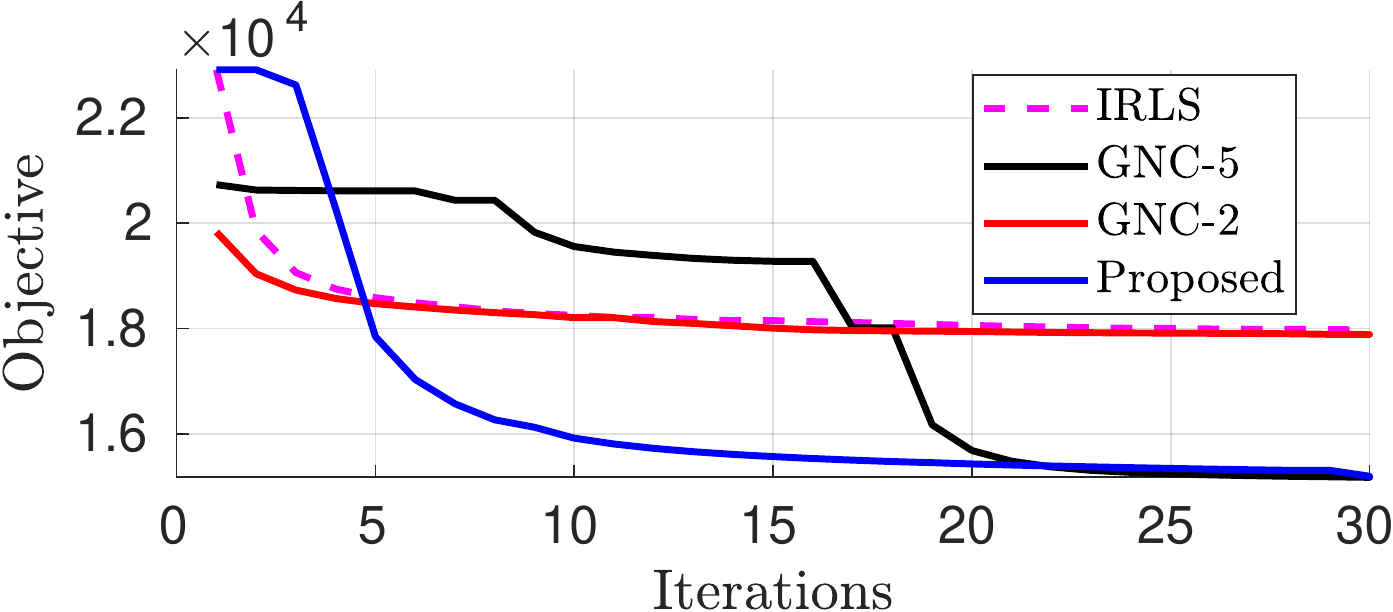}
    \caption{A wrong schedule of GNC may lead to either poor results (GNC-2, which is not better than IRLS) or unnecessary iterations (GNC-5). Here GNC-2 and GNC-5 mean GNC with the number of levels $k$ set to $2$ and $5$, respectively.  Our proposed method provides competitive objective value and converges faster than GNC.}
    \label{fig:gnc_limitation}
\end{figure}
In this section, we briefly review graduated optimization (or graduated non-convexity~\cite{blake1987visual}), which is a popular technique commonly employed to avoid poor local minima in highly non-convex problems.
It also serves as the foundation for our novel method proposed in this work.
Indirectly it is leveraged also in coarse-to-fine schemes used e.g.\ in variational methods for optical flow~\cite{mobahi2012seeing}.
The main idea behind this technique is to optimize the original highly non-convex cost function $\Psi$ by minimizing a sequence of problems $(\Psi^k, \dots, \Psi^0$), where $\kernel^0=\kernel$ and $\kernel^{k+1}$ is ``easier" to optimize than $\kernel^{k}$.
Starting from the original robust kernel $\kernel$ (as defined in~\eqref{eq:robust_mean}), the set of ``easier" problems are obtained by a scaled version of of $\kernel$. In particular, from the original minimization problem with the objective function $\Psi(\btheta)$, each problem $\Psi^{k}$ is constructed with a new kernel $\kernel^{k}$, 
\begin{equation}
    \kernel^k(r) = s^2_k \kernel(r/s_k),
\end{equation}
where the scale parameters are chosen such that $s_{k+1} > s_{k}$ and $s_0 = 1$. Figure~\ref{fig:robust_mean} shows an example of a one dimensional robust mean estimation, where we plot the objective values of the problem with the original kernel and its scaled version (with $s=3$). As can be seen, the scaled kernel results in this case in a problem with no poor local minimum.

To the best of our knowledge, methods that rely on graduated optimization achieve state-of-the-art results for large-scale robust estimation tasks (most importantly, bundle adjustment problems) due to their ability to escape poor local minima. However, in practice it is necessary to define a schedule with a fixed number of levels $k$. This requires some knowledge about the problem so that a proper value for $k$ can be assigned. A large value of $k$ may cause unnecessary iterations, which translates to high running time. On the other hand, setting a low $k$ may not provide sufficient scaling levels for the optimizer to avoid poor solutions (as shown in Figure~\ref{fig:gnc_limitation}). Moreover, in some easy applications, although GNC converges to a lower objective than its competitor (e.g., IRLS), the difference between the converged objectives may be insignificant. In such scenarios, an IRLS solver can provide acceptable results within a few iterations, while it may take longer for a GNC solver to go through all $k$ levels. However, using IRLS poses a risk of converging to bad local minima.
Therefore, there is a trade-off between the selecting a solver and associated hyper-parameters (such as the annealing schedule in GNC) and the resulting efficiency.


%% file: tex/4-approach.tex
\section{Adaptive Kernel Scaling}

In this section, we describe our novel solver for robust parameter estimation that addresses the above weaknesses of GNC.
Our method is motivated by graduated optimization and its ability to avoid poor local minima.
However, unlike previous graduated schemes employing a fixed schedule of kernel scaling, we consider the scale of each residual as a variable, and allow the scales to be jointly optimized with the set of parameters $\btheta$. This leads us to a new formulation for robust estimation, which is a constrained optimization problem and can be written as
\begin{align}
    \min_{\btheta, \{\sigma_i\}} && \sum_{i=1}^N \kernel \left(\frac{\|\br_i (\btheta)\|}{\sigma_i}\right) \qquad \text{s.t. } \sigma_i = 1 \;\; \forall i = 1,\dots,N.
    \label{eq:scaled_robust}
\end{align}
In contrast to e.g.\ graduated optimization, which maintains usually a single smoothness parameter, we introduce a scaling factor $\sigma_i$ for each residual.
Consequently, each scale $\sigma_i$ evolves differently during the optimization process.
Clearly, \eqref{eq:scaled_robust} does not appear helpful, as enforcing the constraints $\sigma_i=1$ strictly (i.e.\ maintaining a feasible solution throughout) makes \eqref{eq:scaled_robust} equivalent to the original task~\eqref{eq:robust_mean}.
Strategies such as graduated optimization do not maintain strictly feasible iterates, but use a schedule for $\sigma_i$ to eventually satisfy the constraints. Turning the original problem~\eqref{eq:robust_mean} into a constrained optimization problem~\eqref{eq:scaled_robust} has two potential benefits: first, a larger set of optimization methods is applicable, and second, intermediate solutions may be infeasible but at the same time correspond to smoother problem instances.

Observe that in order to obtain a solution for~\eqref{eq:scaled_robust}, besides the initialization $\btheta_0$ for the parameters, one can also initialize the scales $\sigma_i$ to values that are greater than $1$ and expect that the solver will drive $\sigma_i$ to the feasible region $\sigma_i=1$ of~\eqref{eq:scaled_robust}. 
Therefore, by considering the problem~\eqref{eq:scaled_robust} and setting $\sigma_i$ to initial values greater than $1$, we are effectively conducting kernel scaling, which provides the potential of escaping poor local minima. In contrast to graduated optimization, the internal workings of the optimization method determine how feasibility of $\sigma_i$ is eventually achieved. In particular, $\sigma_i$ may be updated in non-monotonically and therefore being increased during the iterations of the optimization method. In this work we propose to utilize a filter method to address the constrained problem~\eqref{eq:scaled_robust}, since it is a highly flexible and non-monotone framework for constrained optimization problems.



Another---and possibly more intuitive---way to convert the robust cost~\eqref{eq:robust_mean} is to replicate the residuals and enforcing consistency, e.g.
\begin{align}
    \min_{\btheta, \{\bp_i\}} \sum_i \psi\left( \|\bp_i\| \right) \qquad \text{s.t. } \bp_i = \br_i(\btheta).
\end{align}
Using a filter method in this setting can be shown to be related to additive half-quadratic minimization~\cite{geman1995nonlinear}, and experimentally we found it far inferior compared to using~\eqref{eq:scaled_robust} as starting point.

\section{Optimization with Filter Method}
By introducing the scale variables $\{\sigma_i\}$, we obtained a constrained optimization problem as written in~\eqref{eq:scaled_robust}.
One requirement for the optimization method of choice is, that the limit values of $\sigma_i$ must be $1$ when the algorithm converges.
Moreover, any proposed method for solving~\eqref{eq:scaled_robust} should be competitive with existing second-order solvers for problem instances~\eqref{eq:robust_mean} (such as Ceres~\cite{ceres-solver} and  SSBA~\cite{zach2014robust}). This requirement rules out e.g.~first order methods for constrained programs.


\subsection{Background on Filter Method}

Our technique to solve the constrained program~\eqref{eq:scaled_robust} is inspired by the filter method~\cite{fletcher2002nonlinear}, which was initially developed as an alternative to penalty methods in constrained optimization~\cite{nocedal}.
In order to outline the filter method, let us consider a constrained optimization problem,
\begin{equation}
    \min_{\bx \in \bbR^d} f(\bx), \;\; \text{s.t.} \;\;g_i(\bx) = 0, i = 1\dots c,
    \label{eq:constraint_opt}
\end{equation}
where $f, g_i:\bbR^d \mapsto \bbR$ are continuously differentiable functions, and $c$ is the number of constraints. We also define a function $h(\bx) = \sum_i \|g_i(\bx)\|$ to indicate the constraint violation. Obviously, $h(\bx^*)=0$ iff $\bx^*$ is a feasible solution of~\eqref{eq:constraint_opt}. In classical penalty approaches, the constraint violation is incorporated into the objective with a penalty parameter $\mu$ in order to create a new objective (i.e., $f(\bx) + \mu h(\bx)$). The resulting objective can then be optimized using a suitable local method. Usually, $\mu$ increased monotonically according to a specified schedule to ensure that the solution converges to a feasible region of~\eqref{eq:constraint_opt}. One drawback of such approach is that the initial value of $\mu$ and how it is increased must be carefully tuned. Another practical issue with penalty methods is, that feasibility of the solution is only guaranteed when $\mu\to\infty$ (unless one utilizes an exact but usually non-smooth penalizer $h$~\cite{nocedal}).

\input{tex/algo_filter.tex}

\begin{figure}[ht]
    \centering
    \includegraphics[width = 0.85\columnwidth]{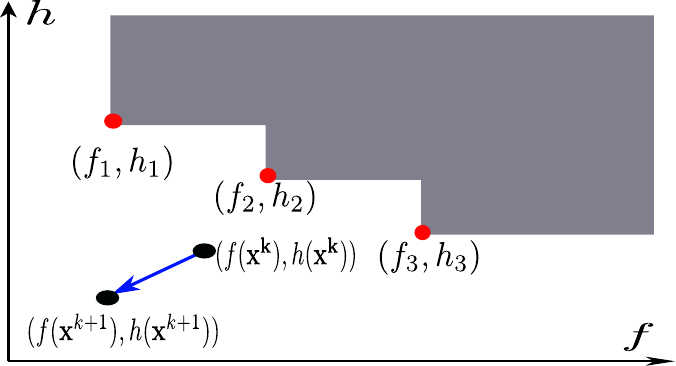}    
    \caption{Example of a filter. The $x$ axis is the objective, while the $y$ axis is the constraint violation. The gray area indicates the forbidden region defined by three mutually non-dominated pairs (shown in red). Optimization with filter method involves finding, from a current $\bx^k$, a new value $\bx^{k+1}$ that is not dominated by the filter. A step that reduces both $f$ and $h$ is preferable (as illustrated by the blue arrow).}
    \label{fig:filter}
\end{figure}

In contrast to penalty methods, Fletcher et al.~\cite{fletcher2002nonlinear} proposes a entirely different mechanism to solve~\eqref{eq:constraint_opt} by introducing the concept of a filter (see Figure~\ref{fig:filter}), which offers more freedom in the step computation. At a current value of $\bx$, let us denote by $F(\bx)$ the pair combining the objective value and the constraint violation, $F(\bx) = (f(\bx), h(\bx)) \in \bbR^2$. For brevity, we sometime use $f$ and $h$ to denote $f(\bx)$ and $h(\bx)$, respectively. 
Given two pairs $F_i = (f_i, h_i)$ and $F_j = (f_j, h_j)$, the concept of \emph{domination} is defined as follows: $F_i$ is said to dominate $F_j$ if $f_i < f_j$ and $h_i < h_j$.  A filter is then defined as a set $\cF = \{F_i\}_{i=1}^m \subseteq \bbR^2$ containing \emph{mutually non-dominating} entries. The filter $\cF$ defines a dominated (and therefore forbidden) region $\bbF$ in the 2D plane. A pair $F_t$ is said to be \emph{accepted by the filter} $\cF$ if it is not dominated by any pair in $\cF$. Figure~\ref{fig:filter} visualizes an example of a filter, where the gray areas is the forbidden region defined by the filter pairs.

Filter methods are iterative, and the basic filter approach is summarized in Algorithm~\ref{alg:filter}.
The filter $\cF$ and the forbidden region $\bbF$ are initialized to empty sets. At the beginning of each iteration, a new pair $(\tdf,\tdh)$ is temporarily added to the filter $\cF$, where $\tdf = f_t - \alpha h_t$ and $\tdh = h_t - \alpha h_t$.
Here $\alpha>0$ specifies the filter margin in order to assure that new points acceptable by the filter must induce a sufficient reduction in the objective value or the constraint violation.
Thus, convergence to feasible solutions is ensured by a such a margin~\cite{ribeiro2008global}.
The procedure to compute $\bx^{t+1}$ (Line~\ref{alg:line:compute_x} of Alg.~\ref{alg:filter}) will be discussed in the following section. Once $\bx^{t+1}$ is obtained, if the objective is reduced, the pair $(\tdf, \tdh)$ is removed from $\cF$, otherwise it is retained in the filter.
For greatest flexibility in computing $\bx^{k+1}$ (and therefore fastest convergence) the filter should contain as few elements as necessary to guarantee convergence to a feasible solution.
On the other hand, adding already feasible iterates to the filter leads to zero margins and is consequently harmful.
New iterates that only certify a sufficient reduction of the constraint violation lead to the temporarily added filter element made permanent. It can be shown~\cite{ribeiro2008global}, that filter elements are always strictly infeasible, but accumulation points are feasible.
The process is repeated until reaching a stationary point of the problem. Interested readers are referred to~\cite{fletcher2002nonlinear,ribeiro2008global} for more detailed information.

\subsection{Application to Robust Estimation}

Our approach to solve~\eqref{eq:scaled_robust} follows closely the steps described in Algorithm~\ref{alg:filter}. However, the main contribution of our work is a novel strategy to compute $\bx^{t+1}$ that is accepted by the filter.
In addition, our method is able to leverage existing non-linear least-squares solvers.

We restrict $\sigma_i$ to be greater or equal to~1, as $\sigma_i \in (0,1)$ will lead to a harder problem than~\eqref{eq:robust_mean}.
Therefore, it is convenient to re-parameterize $\sigma_i$ as $\sigma_i = 1 + s^2_i$ and we can rewrite the problem~\eqref{eq:scaled_robust} as follows
\begin{align}
    \min_{\btheta, \{s_i\}} && \sum_{i=1}^N \kernel \left(\frac{\|\br_i (\btheta)\|}{1 + s^2_i}\right) \qquad \text{s.t. } s_i = 0 \;\; \forall i.
    \label{eq:scaled_robust_s}    
\end{align}{}
In the context of~\eqref{eq:constraint_opt}, let $\bx = [\btheta^T  \; \bs^T]^T$ where $\bs = [s_1 \dots s_n]^T $ is a vector that collects the values of $s_i$.
Finally, the functions $f(\bx)$ and $h(\bx)$ correspond to
\begin{align}
    f(\bx) = \sum_{i=1}^N \kernel \left(\frac{\|\br_i (\btheta)\|}{1 + s^2_i}\right) & &  h(\bx) = \sum_i s^2_i.
\end{align}

\subsubsection{Cooperative Step}
\label{sec:coop_step}
An appealing feature of Algorithm~\ref{alg:filter} is, that it offers a flexible choice of algorithms to perform variable update, as long as $\bx^{t+1}$ is accepted by the filter (i.e., $\bx^{t+1} \notin \bbF$ as described in Line.~\ref{alg:line:compute_x} of Algorithm.~\ref{alg:filter}).
Like filter methods for non-linear constrained minimization there are two possible steps to obtain a new acceptable iterate: the cooperative step described in this section is the main workhorse of the algorithm. It replaces the sequential quadratic program (SQP) used as the main step in filter methods for general non-linear programs~\cite{fletcher2002nonlinear, ribeiro2008global}.
The cooperative step is complemented with a restoration step as a fall-back option, that is described in the following section.

The cooperative step is motivated by the fact that reducing both the main objective and the constraint violation (i.e.,$f(\bx)$ and $h(\bx)$) by a sufficient amount (as induced by the margin parameter $\alpha$) leads to a new solution that is guaranteed to be acceptable by the filter.
We use a second-order approximation of $f$ and $h$ around the current values $\bx^{t}$,
\begin{align}
    f(\bx^t + \Delta \bx) &= f(\bx^t) + \bg_f^T\Delta \bx + \Delta\bx^T\bH_f \Delta \bx, \nonumber\\
    h(\bx^t + \Delta \bx) &= h(\bx^t) + \bg_h^T\Delta \bx + \Delta\bx^T\bH_h \Delta \bx,
\end{align}
where $\bg_f$ and $\bg_h$ are the gradients, while $\bH_f$ and $\bH_h$ are true or approximated Hessian of $f$ and $h$, respectively. Hence, a cooperative update direction $\Delta\bx$ possibly decreasing both $f$ and $h$ is given by~\cite{fliege2009newton}
\begin{align}
    \arg\min_{\Delta\bx} \max\{\Delta f, \Delta h\},
    \label{eq:min_max_direction}
\end{align}
where $\Delta f = \bg_f^T\Delta \bx + \Delta\bx^T\bH_f \Delta \bx$, and $\Delta h = \bg_h^T\Delta \bx + \Delta\bx^T\bH_h \Delta \bx$. This is a convex quadratic program, which can be efficiently solved using any iterative solver. However, as previously discussed, our ultimate goal is to integrate our algorithm into existing solvers, following~\cite{zach2019pareto} we relax the problem~\eqref{eq:min_max_direction} to
, rather than solving, we aim to find $\Delta\bx^t$ that solves
\begin{align}
    \Delta\bx^t &= \arg\min_{\Delta\bx} \max\{\Delta f, \beta\Delta h\} \nonumber \\
    &= \arg\min_{\Delta\bx} \mu_f \Delta f + \mu_h \Delta h,
    \label{eq:min_max_direction_relaxed}
\end{align}
where $\mu_f >0 $ and $\mu_h > 0$ with $\mu_f + \mu_h = 1$ are suitably chosen coefficients.
$\beta > 0$ is a scaling factor between the objectives that is implicitly determined by solving for $\Delta\bx$. Adding a Levenberg-Marquardt-type damping~\cite{more1978levenberg} with parameter $\lambda$ yields
\begin{align}
    \Delta\bx^{t}  = (\mu_f\bH_f + \mu_h \bH_h + \lambda \bI)^{-1} (\mu_f\bg_f + \mu_h \bg_h).
    \label{eq:step_computation}
\end{align}
If the new iterate $\bx^{t+1} = \bx^t + \Delta\bx^t$ is acceptable by $\cF$, then $\lambda$ is decreased, otherwise increased. We refer to the supplementary material for further details.

With an appropriate choice of $\mu_f$,  $\mu_g$ and a sufficiently large $\lambda$, it can be shown that $\Delta\bx^t$ leads to a reduction of both $f$ and $g$ as long as $\bg_f$ and $\bg_f$ are not pointing in opposite directions~\cite{zach2019pareto}.
If $\bx^t+\Delta\bx^t$ leads to a sufficient decrease of both $f$ and $h$, then this new solution is by construction acceptable by the current filter.
Otherwise, the new iterate may be still acceptable, but increases either $f$ or $h$ (and is therefore a non-monotone step).
If the new solution is not acceptable by the filter, then a non-monotone restoration step is applied (that also leads to an increase of either $f$ or $h$).
The filter condition ensures that $h$ eventually converges to~0. 

\subsubsection{Restoration Step}
\label{sec:restore_step}
Although~\eqref{eq:step_computation} gives us a way to compute preferable update step, it does not guarantee to provide always steps that are accepted by the filter.
In such cases, we revert to a restoration step described below. 

In the filter methods literature a restoration step essentially reduces the constraint violation and is applied if the SQP step did not yield an acceptable new iterate.
Note that in our setting, just reducing the constraint violation is trivial, and a perfectly feasible solution can be obtained by setting $s_i=0$ for all $i$.
A good restoration step aims to yield a good starting point for the next main step (which is SQP in traditional filter methods and a cooperative step in our approach).
Consequently, the goal of our restoration step is to determine a suitable new solution for the subsequent cooperative step.
One simple criterion for such a new point is given by the angle between the gradients of $f$ and $h$, which is to be minimized in order to facilitate cooperative minimization.
Our deliberate design choice is to adjust only the parameters $s_i$ in the restoration step, i.e.
\begin{align}
    \Delta\bx &= \gamma \binom{0}{\Delta\bs},
\end{align}{}
where $\gamma$ is a step-size determined by a grid search,
\begin{equation}
    \gamma = \arg\min_{\gamma} \angle (\bg_f(\bx + \Delta\bx), \bg_h(\bx + \Delta\bx)).
\end{equation}
Note that adjusting $\bs$ affects both $\bg_f$ and $\bg_h$.
The search direction $\Delta\bs$ is chosen as $\Delta\bs=-\bs$.
Due to the particular choice of $h$ this search direction coincides with the direction to the global minimum $\bs =0$ of $h$, with the negated gradient $-\nabla_{\bs} h(\bs)$, and with a Newton step optimizing $h$.
We limit the search for $\gamma$ to the range $[-1/2,1/2]$. A detailed summarization of our algorithm is provided in the supplementary material.



%% file: tex/algo_filter.tex
\begin{algorithm}[ht]\centering
\caption{Optimization with Filter Method}
\label{alg:filter}                         
\begin{algorithmic}[1]                   
	\REQUIRE Initial solution $\bx^0$, filter margin $\alpha$, \texttt{max\_iter}
	\STATE  Initialization: $t \leftarrow 0$, $\cF \leftarrow \emptyset$ , $\bbF \leftarrow \emptyset$
	\WHILE{\texttt{true} and $t<$ \texttt{max\_iter}}
	    \IF{$\bx^t$ is stationary}
	        \STATE break;
	    \ENDIF
	    \STATE $\tilde{f}\leftarrow f_t - \alpha h_t$;  $\tilde{h} \leftarrow h_t - \alpha h_t$
	    \STATE $\cF \leftarrow \cF \cup \{(\tilde{f}, \tilde{h})\}$
	    \STATE $\bF_{t+1} \leftarrow \{\bx | f(\bx) \ge \tilde{f}, h(\bx) \ge \tilde{h} \}$ 
	    \STATE $\bbF \leftarrow \bbF \cup \bF_{t+1}$
	    \STATE Compute $\bx^{t+1} \notin \bbF$ \label{alg:line:compute_x} (Sec.~\ref{sec:coop_step} and~\ref{sec:restore_step})
	    \IF{$f(\bx^{t+1}) < f(\bx^{t})$}
	        \STATE $\cF \leftarrow \cF \setminus \{(\tilde{f}, \tilde{h})\}$; $\bbF \leftarrow \bbF \setminus \bF_{t+1}$
	    \ENDIF
	    \STATE $t \leftarrow t + 1$
	\ENDWHILE
	\RETURN $\bx^t$
\end{algorithmic}
\end{algorithm}

%% file: tex/5-results.tex
\section{Experimental Results}
\input{tex/fig_convergence.tex}
In this section, we evaluate the performance of our proposed algorithm and compare it against current state-of-the-art algorithms for large-scale robust estimation, including: IRLS~\cite{green1984iteratively}, Multiplicative Half-Quadratic Lifting (M-HQ)~\cite{zach2014robust}, Graduated Non-Convexity (GNC) as implemented in~\cite{zach2018descending}, and LM-MOO~\cite{zach2019pareto}. For brevity, we name our method ASKER (which stands for \textbf{A}daptive \textbf{S}caling of \textbf{KER}nels). Following recent works~\cite{zach2019pareto,  zach2018descending}, we use large-scale bundle adjustment as the main problem to test our method, and  a small section of the experiments is devoted to evaluate the stereo dense correspondences problem.

We implement our algorithm in C++ based on the framework provided by SSBA\footnote{\url{https://github.com/chzach/SSBA}}, which is built on direct sparse linear solvers\footnote{\url{http://www.suitesparse.com}}. All experiments are executed on an Ubuntu workstation with an AMD Ryzen 2950X CPU and 64GB RAM. Other methods are also implemented based on the SSBA framework. For better visualization of the figures, we only compare our algorithm against the methods listed above, which are the best representatives for baseline and state-of-the-art approaches. As reported in~\cite{zach2018descending}, methods such as square-rooting the kernels~\cite{engels2006bundle} or Triggs correction~\cite{triggs1999bundle} do not offer  much improvement compared to IRLS, hence we omit these in our experiments. All methods are initialized from the same starting point. In the following, we report the main quantitative and qualitative results, more results on bundle adjustment and reconstructed structures are provided in the supplementary material.

\begin{figure*}[htb]
    \centering
    \includegraphics[width=0.48\textwidth]{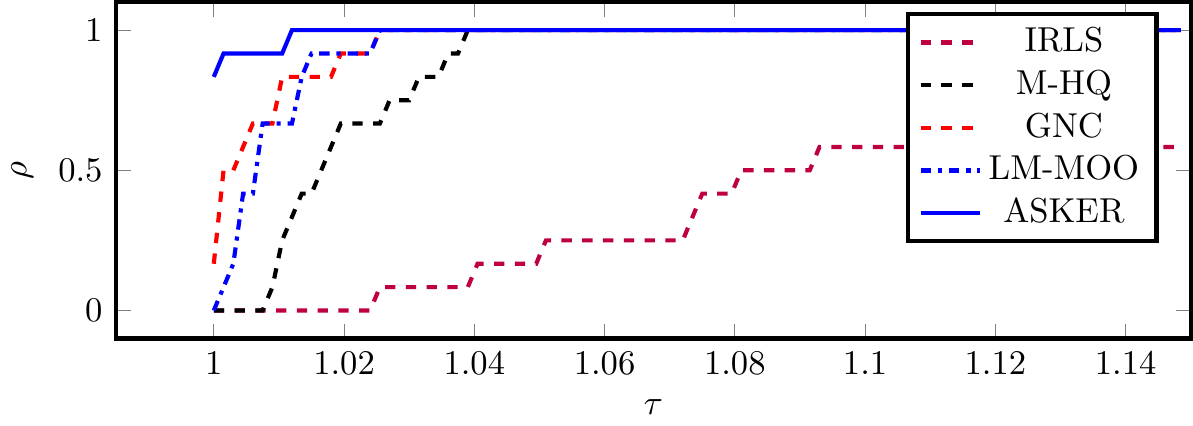}
    \includegraphics[width=0.48\textwidth]{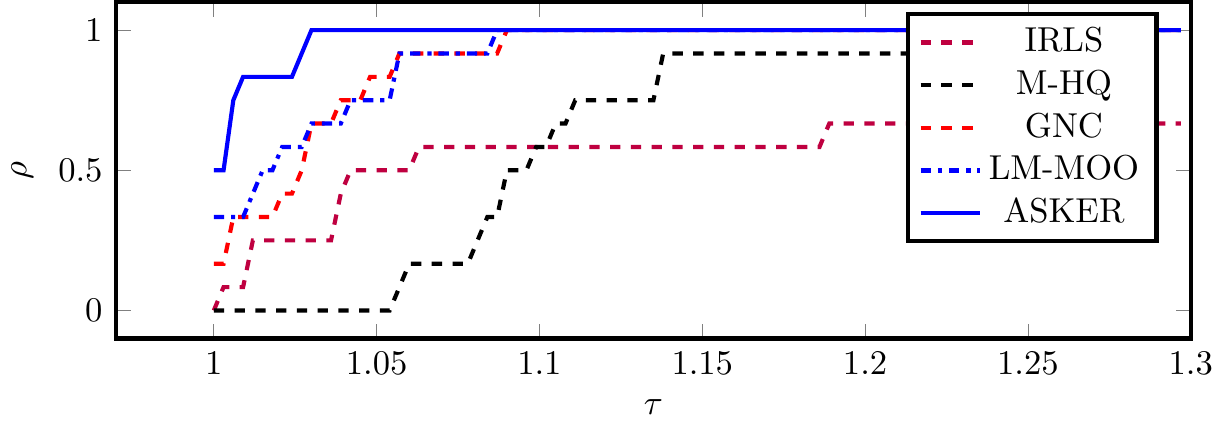}
    \caption{Performance profiles of the best cost (left) and average cost (right) after 100 iterations corresponding to the detailed results in Fig.~\ref{fig:results_convergence}.
    The average cost can be seen as ``area-under-the-curve'' in Fig.~\ref{fig:results_convergence} and is one measure on how fast the target objective decreases w.r.t.\ the iterations.}
    \label{fig:perf_profiles}
\end{figure*}

\subsection{Robust Bundle Adjustment}
We use the well-known dataset provided by~\cite{agarwal2010bundle} for our robust bundle adjustment experiments\footnote{The datasets can be downloaded at~\url{https://grail.cs.washington.edu/projects/bal/}}. This dataset contains the 3D structures reconstructed from a set of images as described in~\cite{agarwal2010bundle}. The whole reconstruction is divided into five sub-datasets: Ladybug, Trafalgar Square, Dubrovnik, Venice, and Final. We extract 20 sequences (the list is provided in the supplementary material), which are considered challenging for robust estimation and have been used throughout recent works~\cite{zach2019pareto, zach2018descending}. We conduct metric bundle adjustment that optimizes the camera poses and the 3D points, with the residual function as described in~\eqref{eq:bundle_residual}. In the following, due to space limit, we only report results for 12 representative datasets. The rest of the results can be found in the supplementary material.

We investigate the performance of the algorithms by executing all the methods with a maximum number of $100$ iterations. The values of $(\mu_f, \mu_h)$ are set to $(0.7, 0.3)$ respectively. The filter margin $\alpha$ is set to $10^{-4}$, and all $s_i$ are initialized to $5.0$ for all datasets. 

Fig.~\ref{fig:results_convergence} depicts the evolution of best encountered objective values with respect to the run time (in seconds) for the participating methods.
The first message that can be extracted from this figure is, that the final objective values obtained by our method are similar or lower that the best ones found by GNC or LM-MOO. Classical IRLS, as anticipated, is suffering from poor local minima in most problem instances (refer to the supplementary material for visual results of the reconstructed structures, where we show that ASKER provides much better structures compared to the poor solutions obtained by IRLS). M-HQ provides better results than IRLS, but is generally not as competitive as GNC, LM-MOO and ours.


Fig.~\ref{fig:perf_profiles} summarizes the findings illustrated in Fig.~\ref{fig:results_convergence} using performance profiles~\cite{dolan2002benchmarking}. A performance profile indicates for each method the fraction $\rho$ of problem instances, for which the method is within a factor $\tau$ compared to the best method (with respect to a chosen performance measure). In Fig.~\ref{fig:perf_profiles}~(left) the performance profile w.r.t.\ to the best objective value reached after 100 iterations is shown.
In several applications a fast decrease of the objective value is of interest, e.g.\ in real-time scenarios when the solver can be interrupted at any time.
We use the mean objective value averaged over the first 100 iterations as rough indicator of how fast a method decreases the objective (which can be also understood as the area under the respective graph in Fig.~\ref{fig:results_convergence}).
The resulting performance profile is shown in Fig.~\ref{fig:perf_profiles}~(right).
The take-home message from these figures is, that our proposed method (ASKER) is leading in both profiles, i.e.\ yields very competitive local minima and good convergence behavior in most of the instances.

Table~\ref{table:results_inliers} shows the inlier fractions (with the inlier threshold set to $1$ pixel) obtained by the algorithms for some datasets. This table conforms with the performance showed in Figure~\ref{fig:results_convergence}, where our algorithm achieved competitive results compared to GNC and LM-MOO.

\input{tex/table_inliers.tex}

\subsection{Dense Correspondences}
\begin{figure}
    \centering
    \includegraphics[width = 0.32\columnwidth]{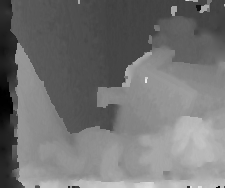}
    \includegraphics[width = 0.32\columnwidth]{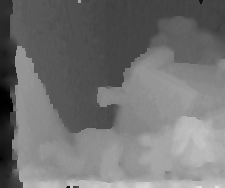}
    \includegraphics[width = 0.32\columnwidth]{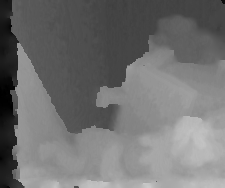}
    
    \includegraphics[width = 0.32\columnwidth]{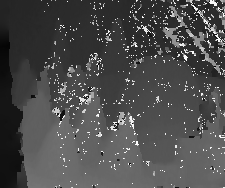}
    \includegraphics[width = 0.32\columnwidth]{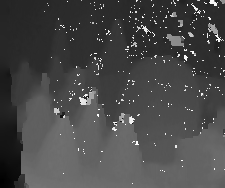}
    \includegraphics[width = 0.32\columnwidth]{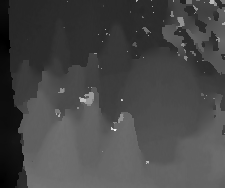}
    
    \caption{Visual results of dense correspondences experiment for the Teddy (top) and Cones (bottom) image pairs. From left to right: LM-MOO, HOM, and Ours.}
    \label{fig:visual_dense}
\end{figure}

\begin{figure}
    \centering
    \includegraphics[width = 0.8\columnwidth]{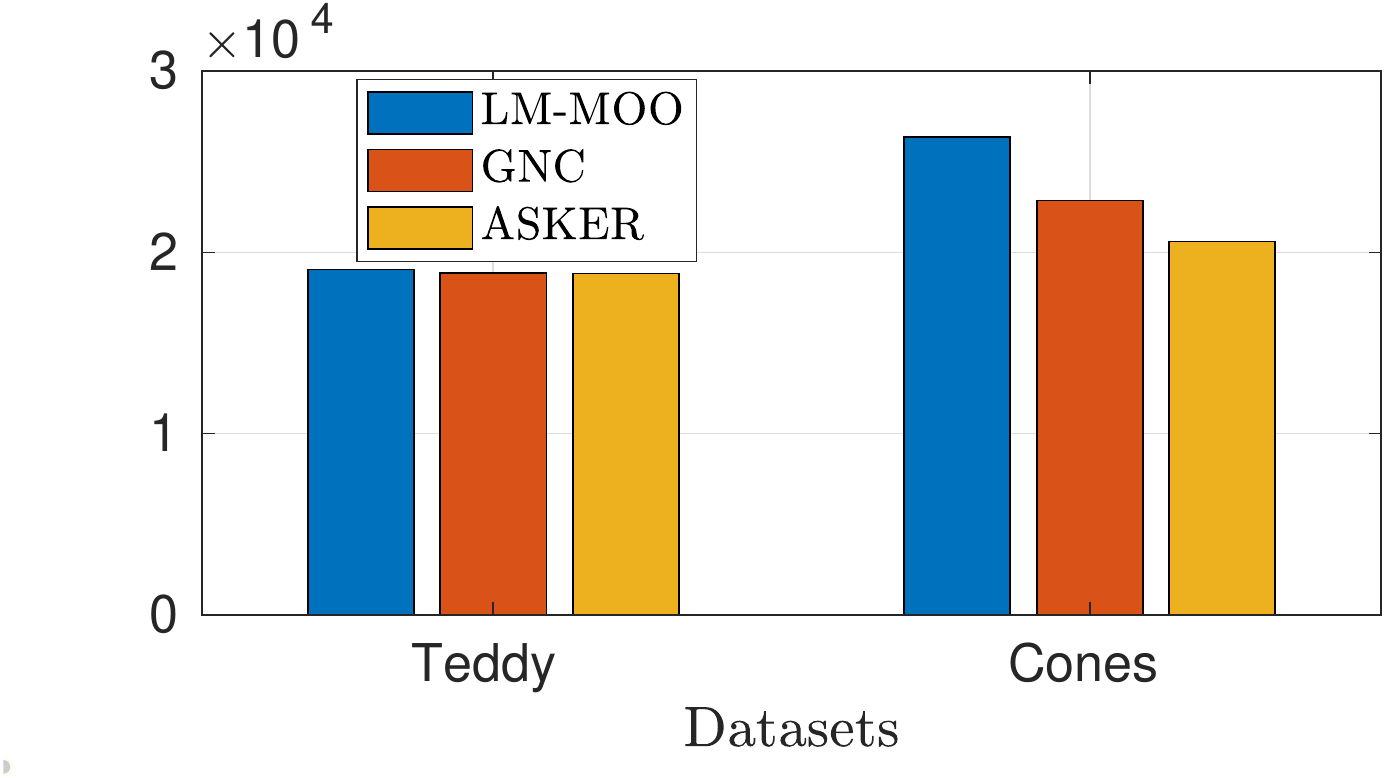}
    \caption{Final objective obtained by the methods for dense correspondences.}
    \label{fig:objective_stereo}
\end{figure}

Following~\cite{zach2019pareto}, we also test our algorithm on the stereo dense correspondence problem, which is also considered a challenging problem in robust estimation. The robust objective can be written as~\cite{zach2019pareto},
\begin{align}
    \sum_{p\in \mathcal{V}} \left(\eta \sum_{k=1}^K \kernel_{\text{data}} (d_p - \hat{d}_{p,k}) + \sum_{q\in \mathcal{N}(p)} \kernel_{\text{reg}}(d_p - d_q)\right),
\end{align}
where $v$ is is a pixel in the image $\mathcal{V}$, $\mathcal N(p)$ is the 4-neighborhood, and $\hat{d}_{p,k}$ is the position of the $k$ local minimum. The parameter $\eta$ is set to~4. The parameter settings of our method follows the bundle adjustment experiments, and all methods are executed with a maximum of $150$ iterations. We also compare our algorithm against GNC and LM-MOO in this experiment. Figure~\ref{fig:visual_dense} shows the visual results, while Figure~\ref{fig:objective_stereo} plots the objectives obtained by the methods. Observe that we also achieve competitive results for this problem, where our method offers lower objectives than LM-MOO and GNC and provides visually better depth map estimations.

%% file: tex/fig_convergence.tex
\begin{figure*}[ht]
    \centering
    \begin{subfigure}[b]{0.24\linewidth}
        \includegraphics[width=\textwidth]{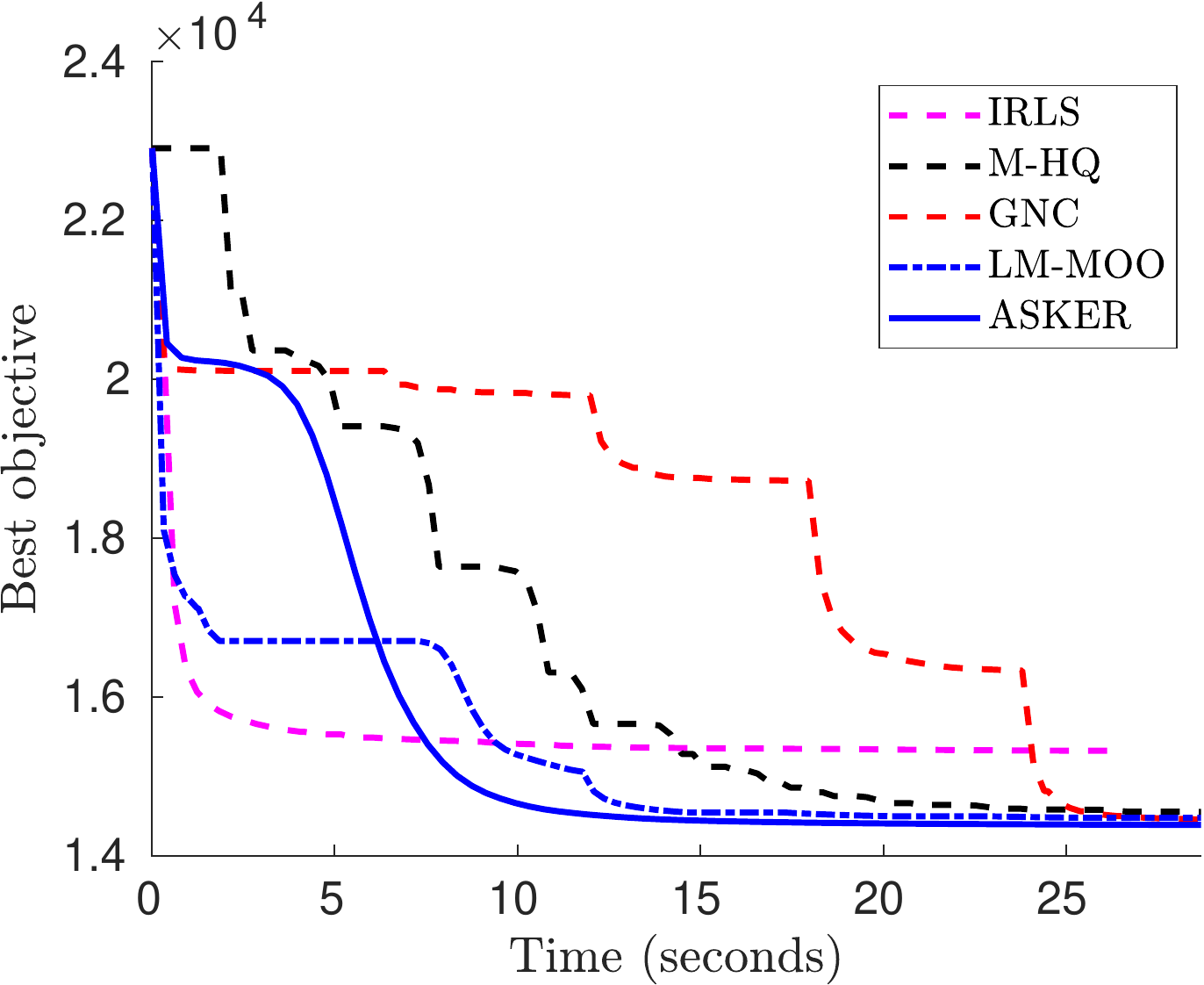}
        \caption{Trafalgar-126}
        \label{subfig:tra126}
    \end{subfigure}
    \begin{subfigure}[b]{0.24\linewidth}
        \includegraphics[width=\textwidth]{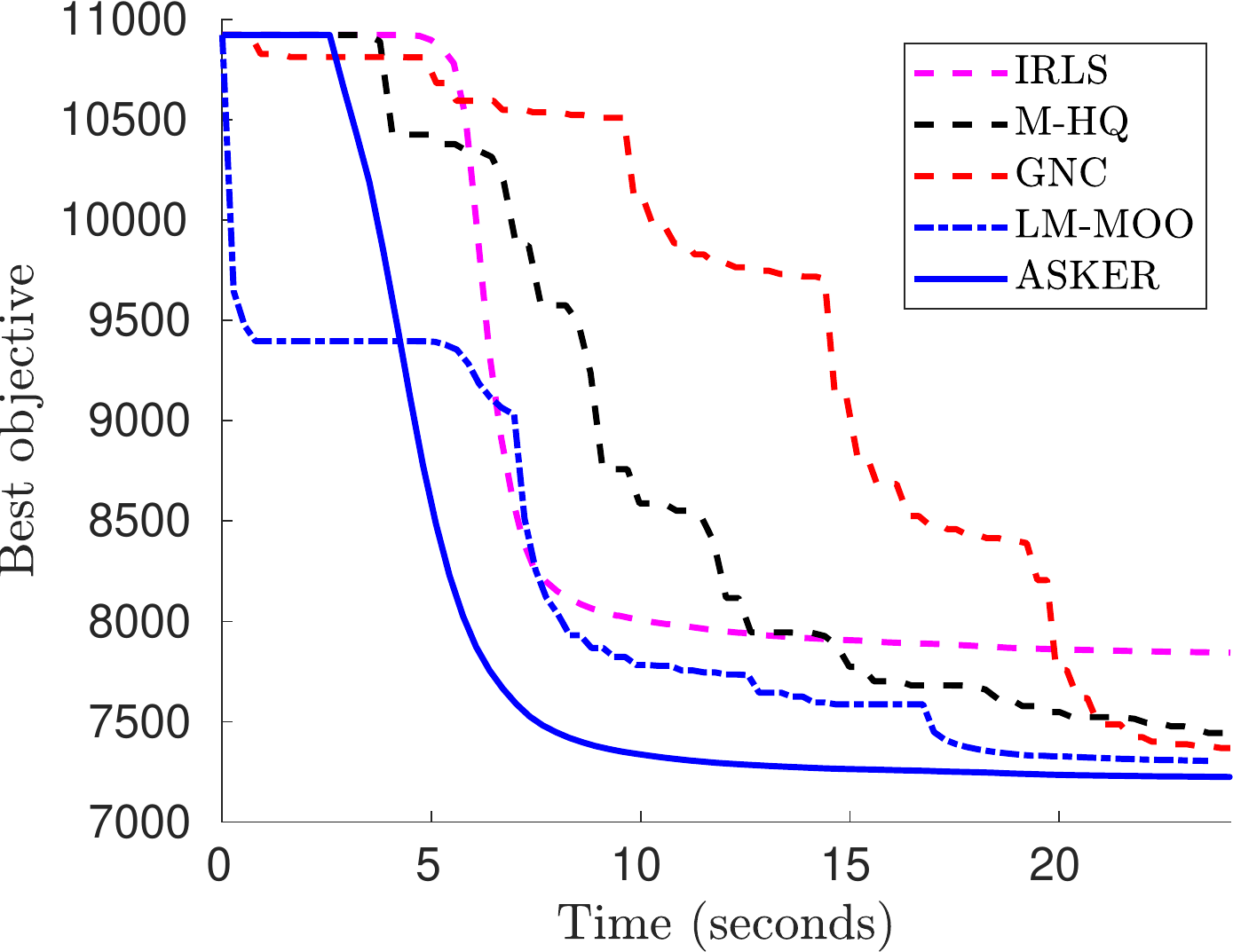}
        \caption{Trafalgar-138}
        \label{subfig:tra138}
    \end{subfigure}
    \begin{subfigure}[b]{0.24\linewidth}
        \includegraphics[width=\textwidth]{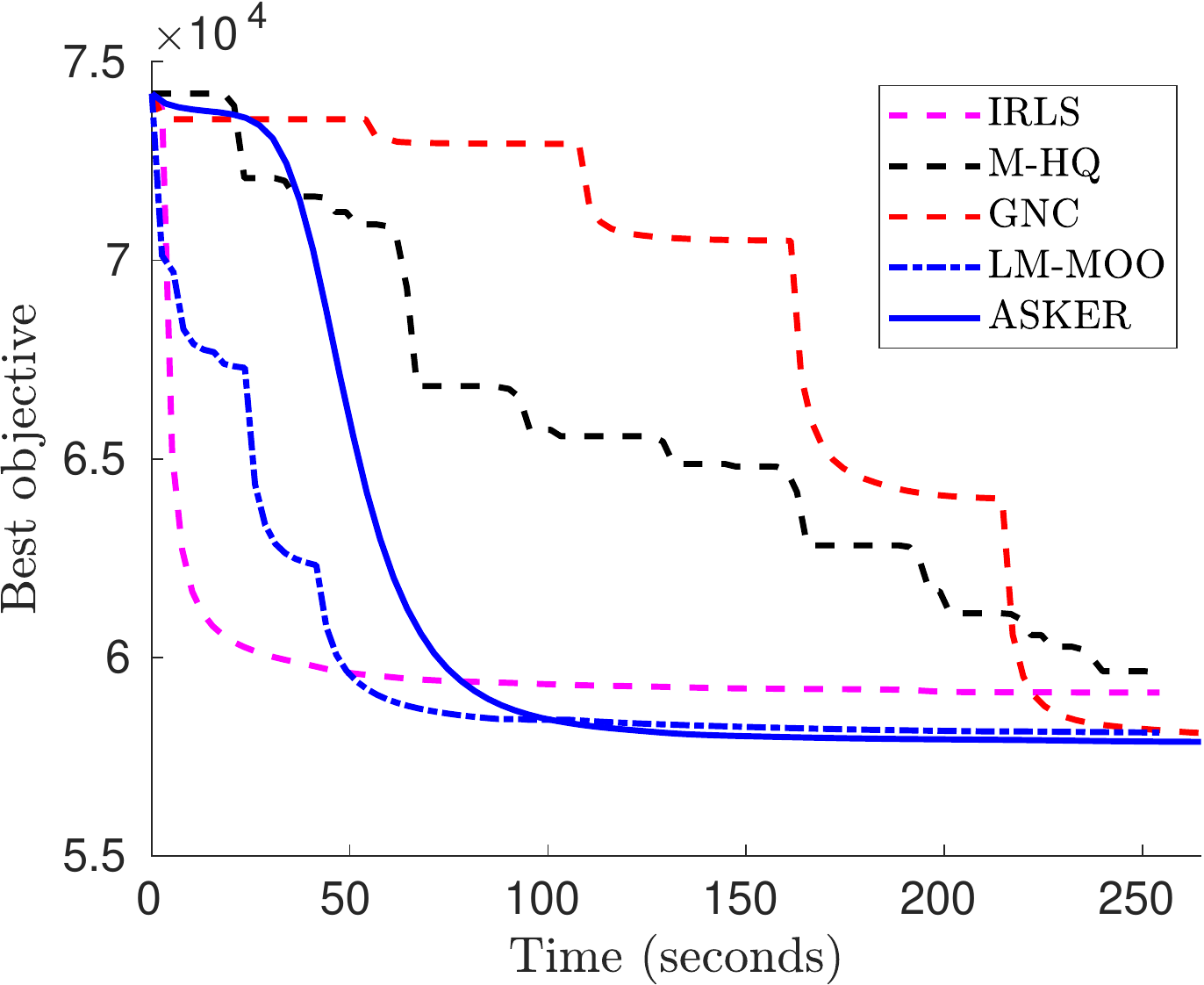}
        \caption{Dubrovnik-150}
        \label{subfig:dub150}
    \end{subfigure}
    \begin{subfigure}[b]{0.24\linewidth}
        \includegraphics[width=\textwidth]{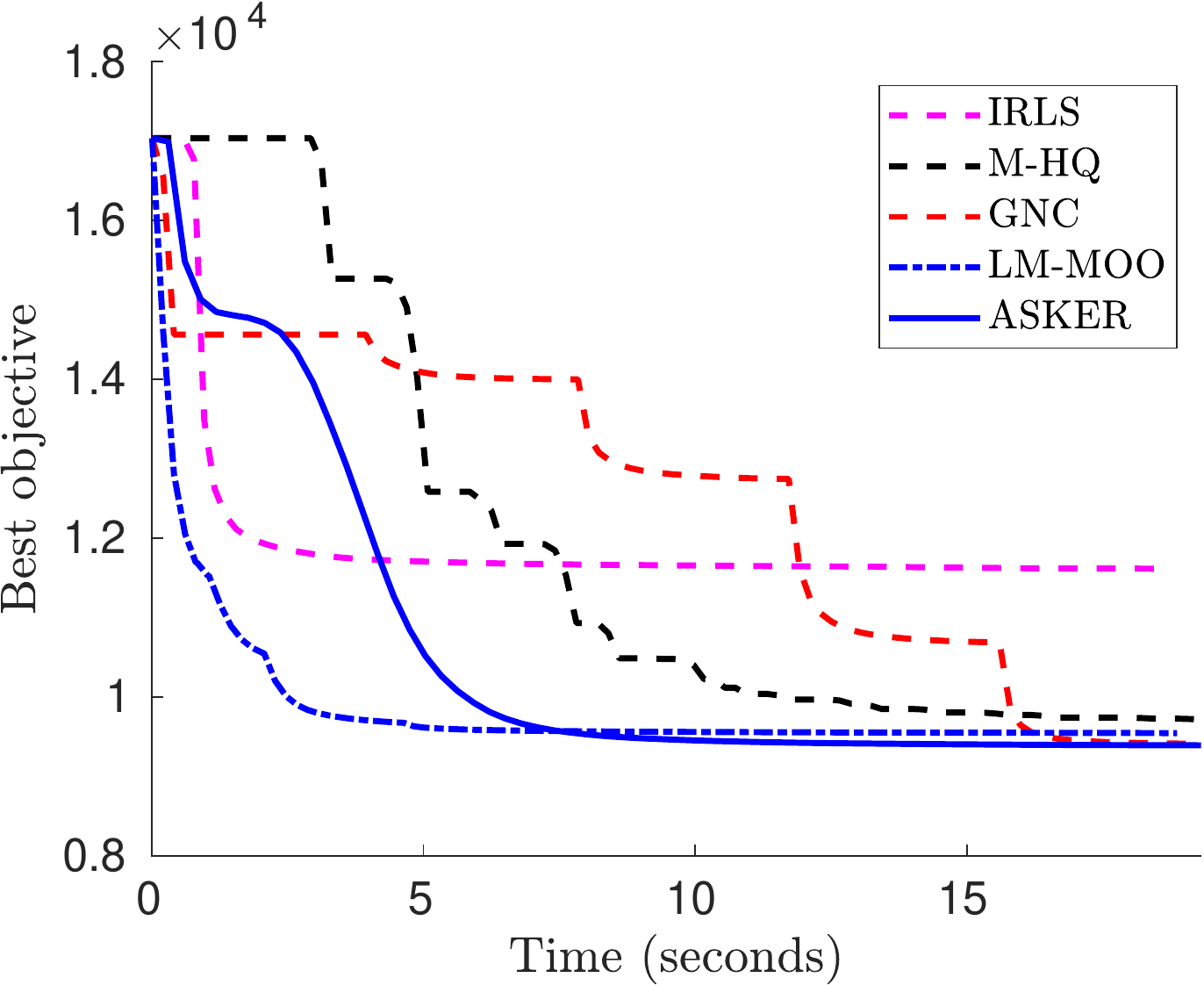}
        \caption{Dubrovnik-16}
        \label{subfig:dub16}
    \end{subfigure}
    
    \begin{subfigure}[b]{0.24\linewidth}
        \includegraphics[width=\textwidth]{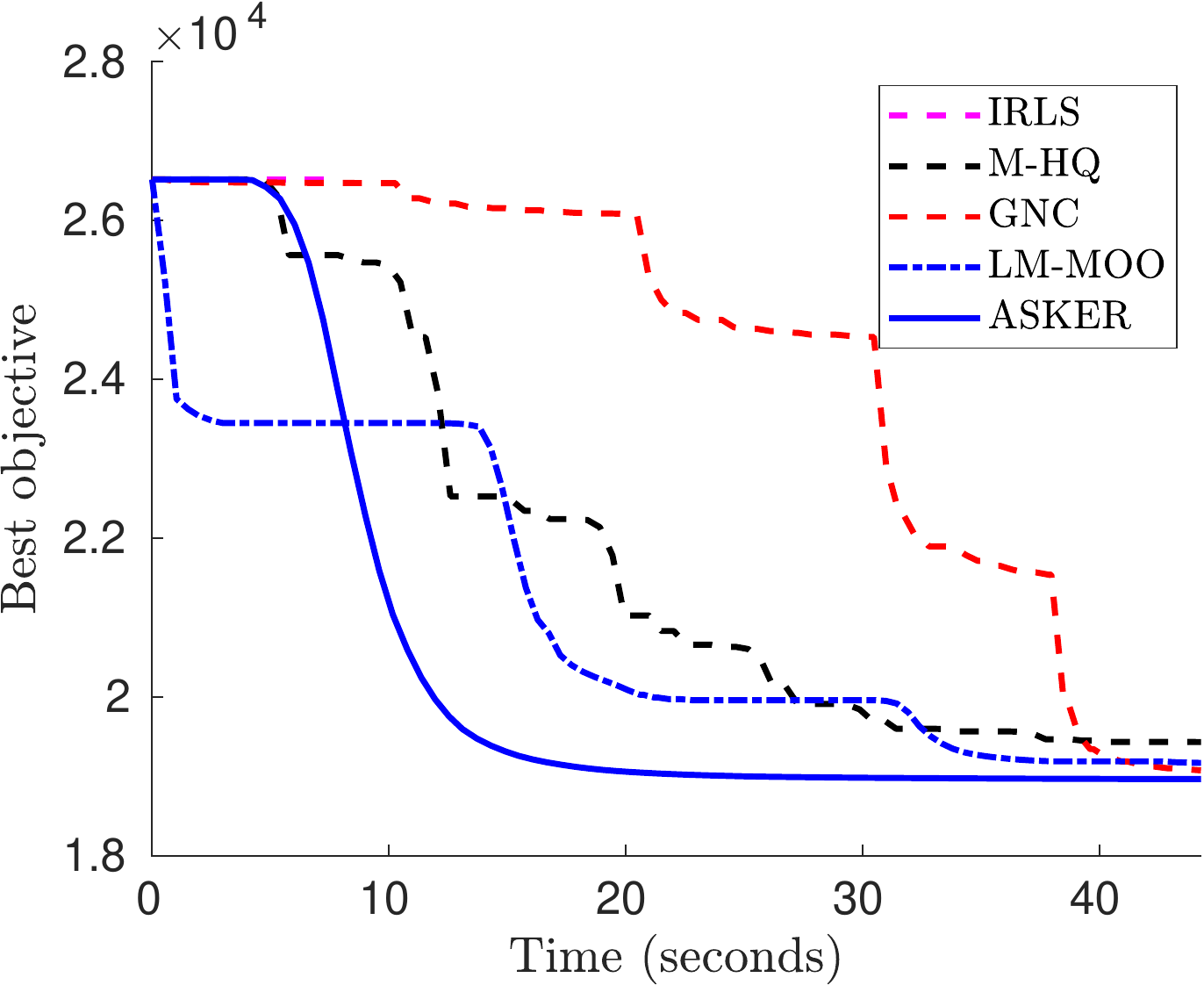}
        \caption{Trafalgar-201}
        \label{subfig:tra201}
    \end{subfigure}
    \begin{subfigure}[b]{0.24\linewidth}
        \includegraphics[width=\textwidth]{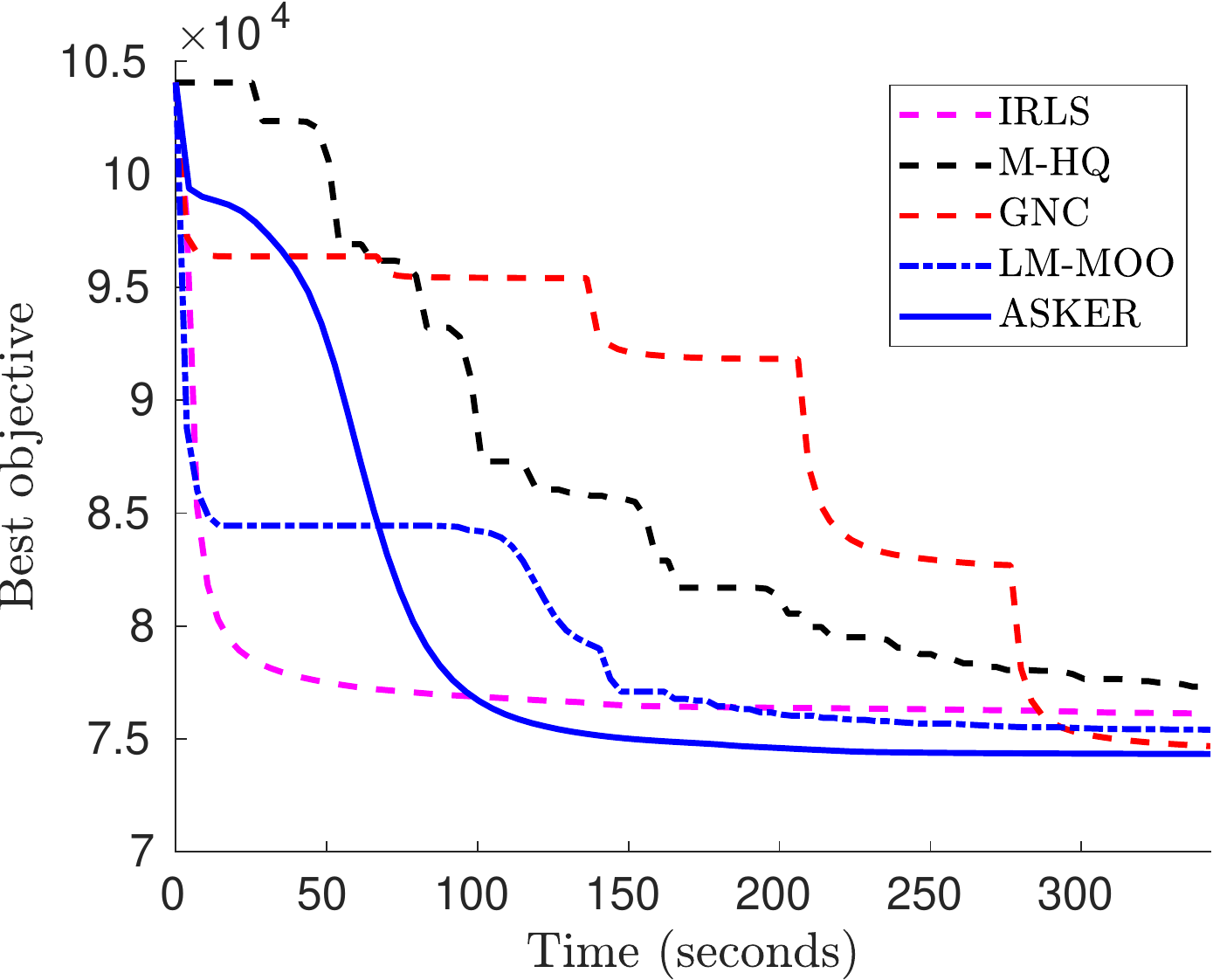}
        \caption{Dubrovnik-202}
        \label{subfig:dub202}
    \end{subfigure}
    \begin{subfigure}[b]{0.24\linewidth}
        \includegraphics[width=\textwidth]{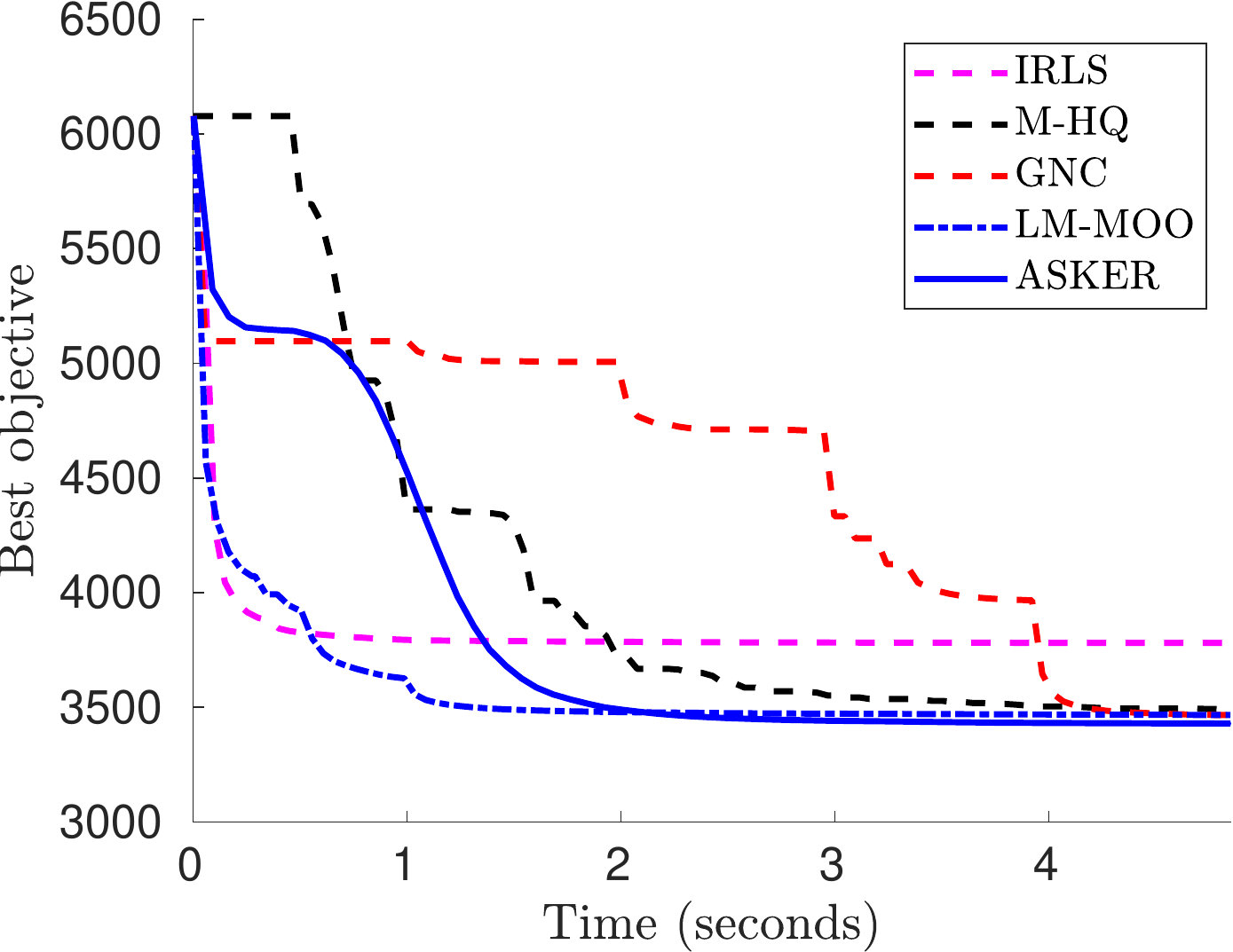}
        \caption{Trafalgar-21}
        \label{subfig:tra21}
    \end{subfigure}
    \begin{subfigure}[b]{0.24\linewidth}
        \includegraphics[width=\textwidth]{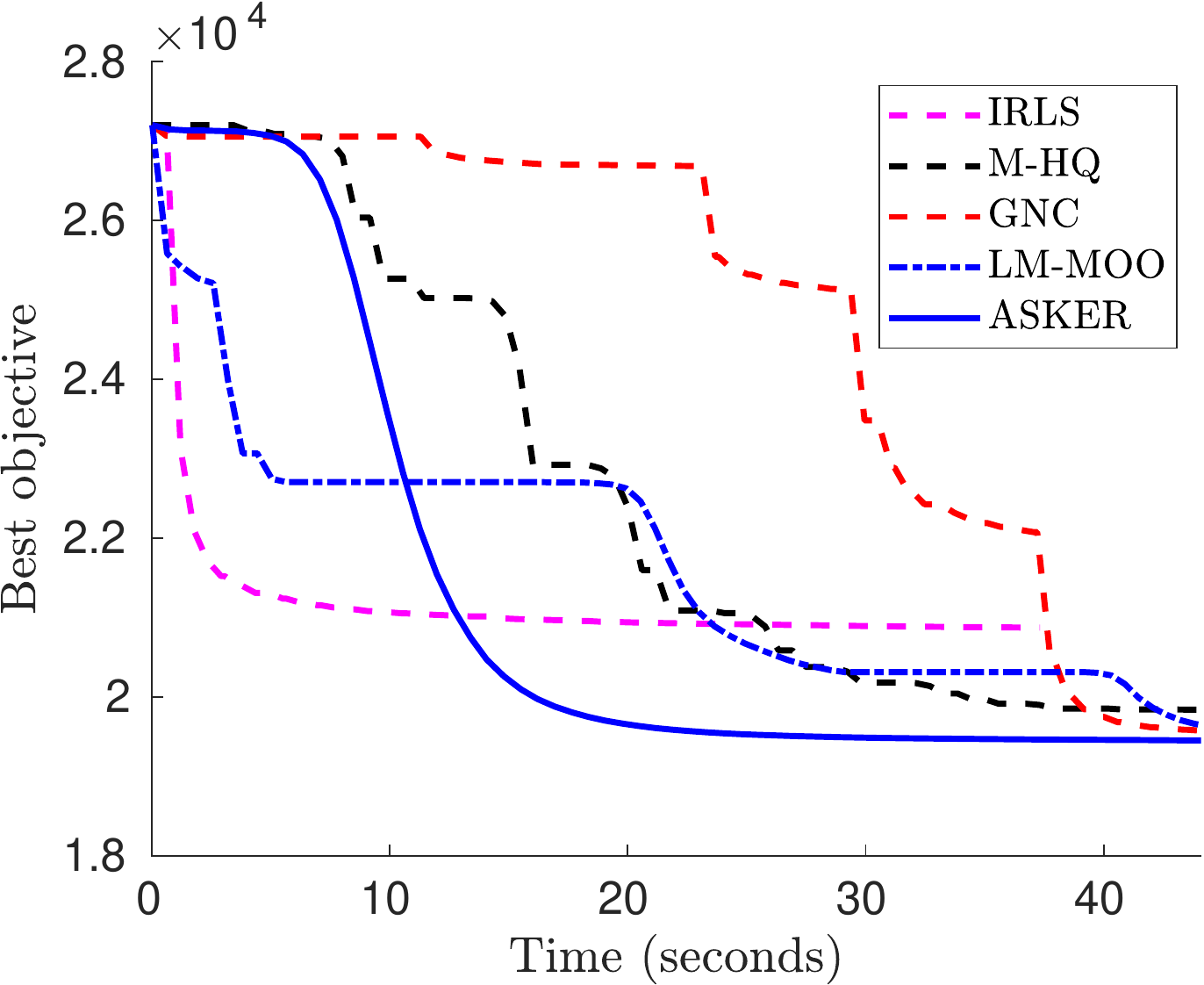}
        \caption{Trafalgar-225}
        \label{subfig:tra225}
    \end{subfigure}
    
    \begin{subfigure}[b]{0.24\linewidth}
        \includegraphics[width=\textwidth]{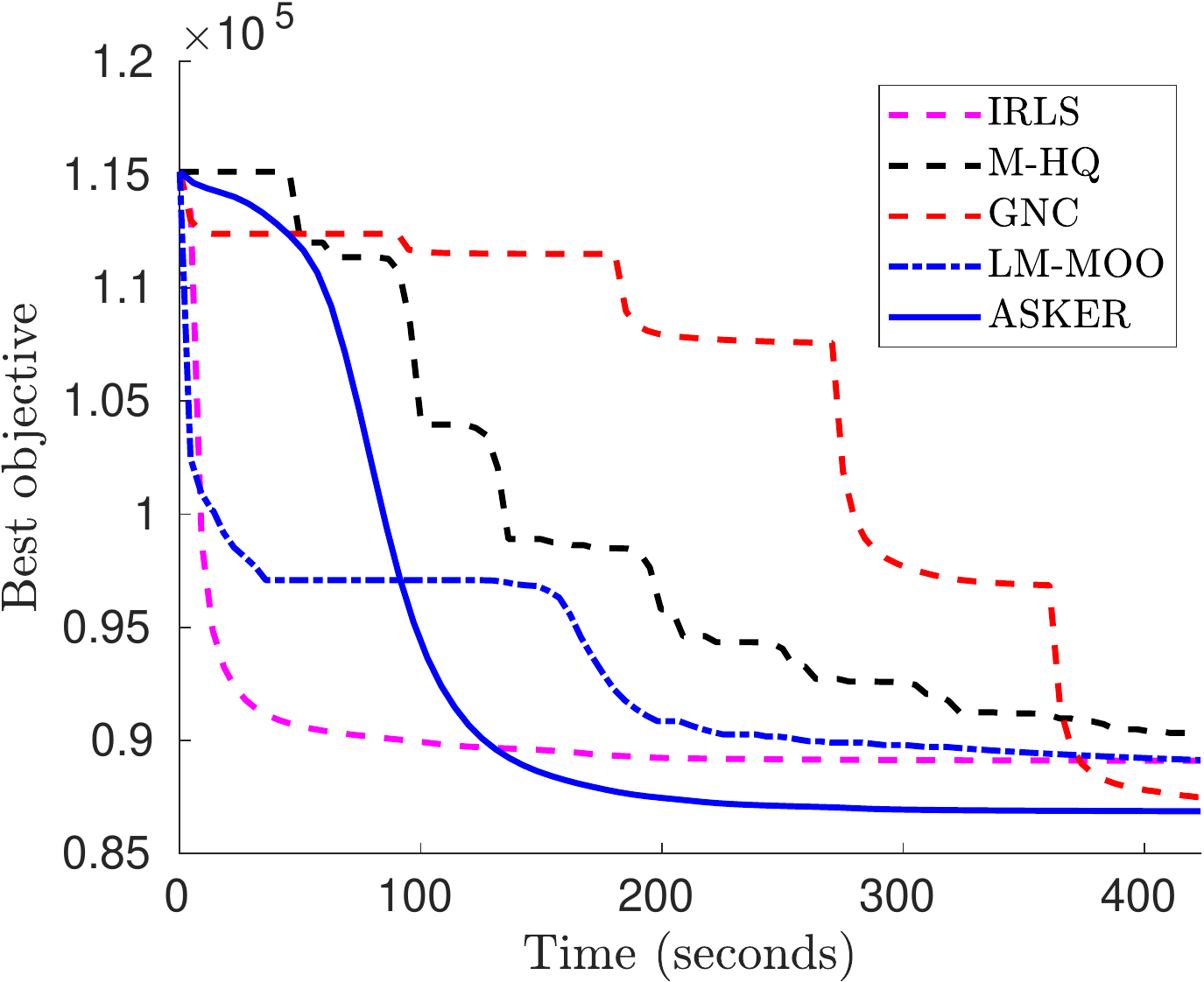}
        \caption{Dubrovnik-253}
        \label{subfig:dub253}
    \end{subfigure}
    \begin{subfigure}[b]{0.24\linewidth}
        \includegraphics[width=\textwidth]{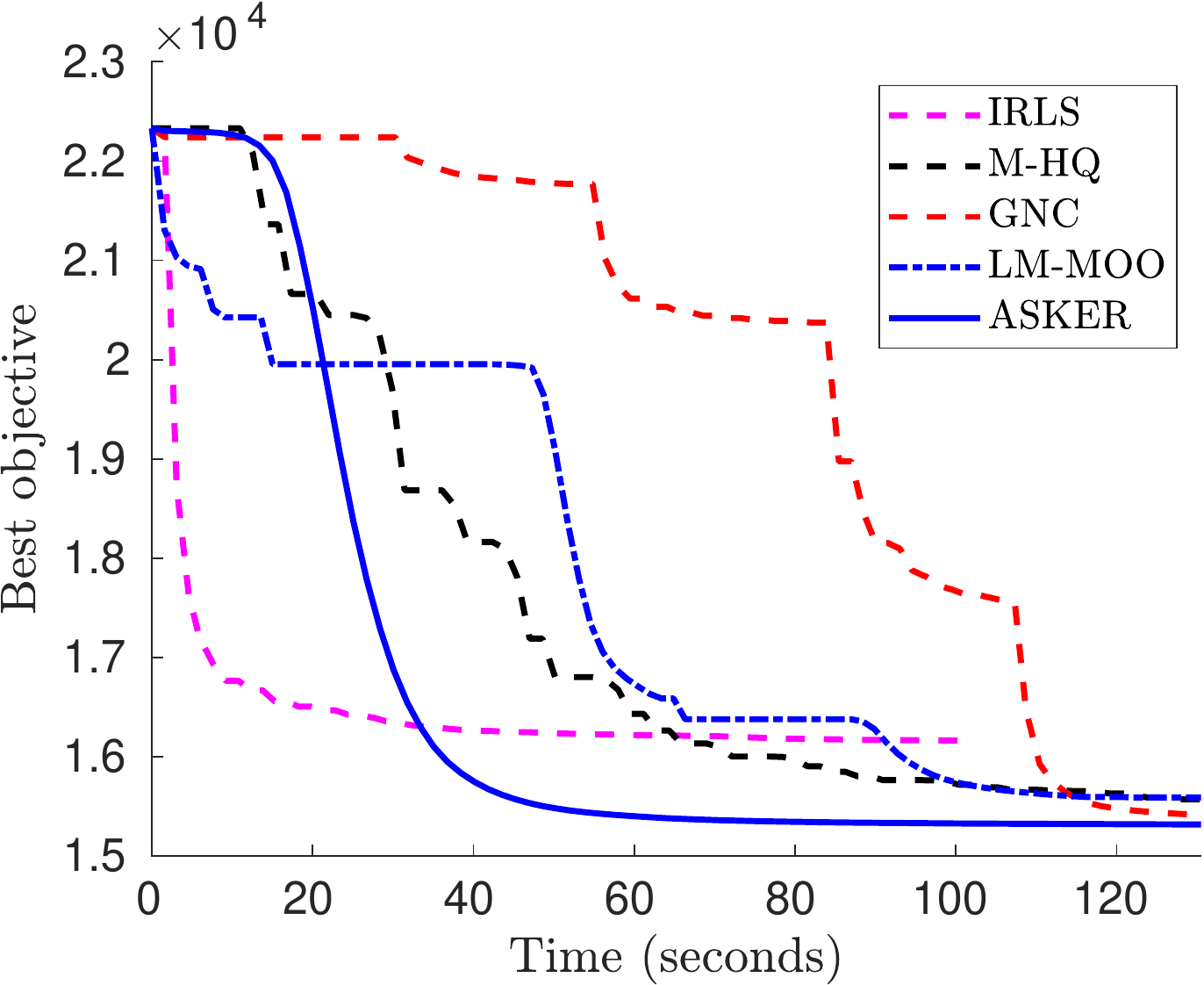}
        \caption{Ladybug-318}
        \label{subfig:lad318}
    \end{subfigure}
    \begin{subfigure}[b]{0.24\linewidth}
        \includegraphics[width=\textwidth]{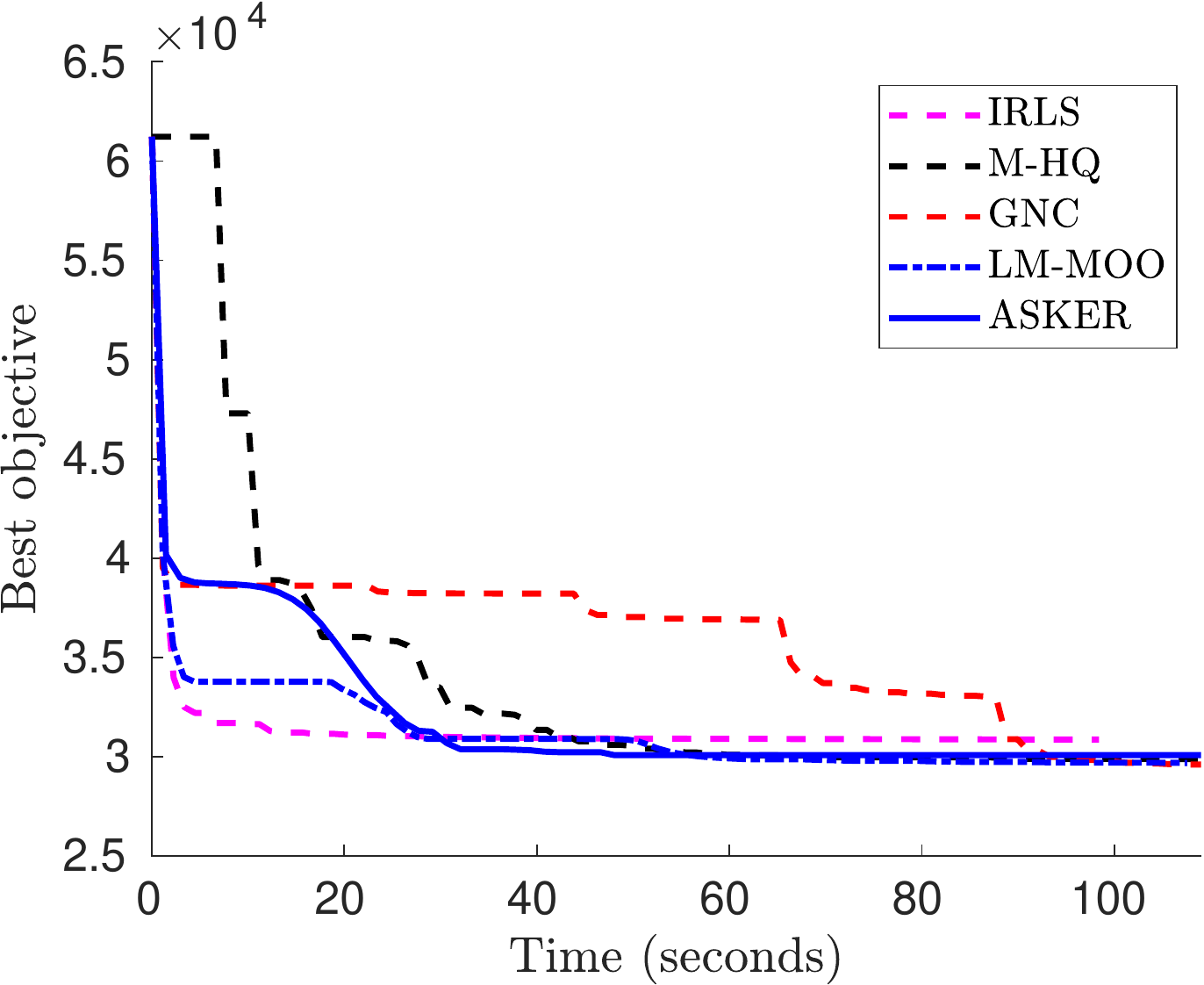}
        \caption{Final-93}
        \label{subfig:fin93}
    \end{subfigure}
    \begin{subfigure}[b]{0.24\linewidth}
        \includegraphics[width=\textwidth]{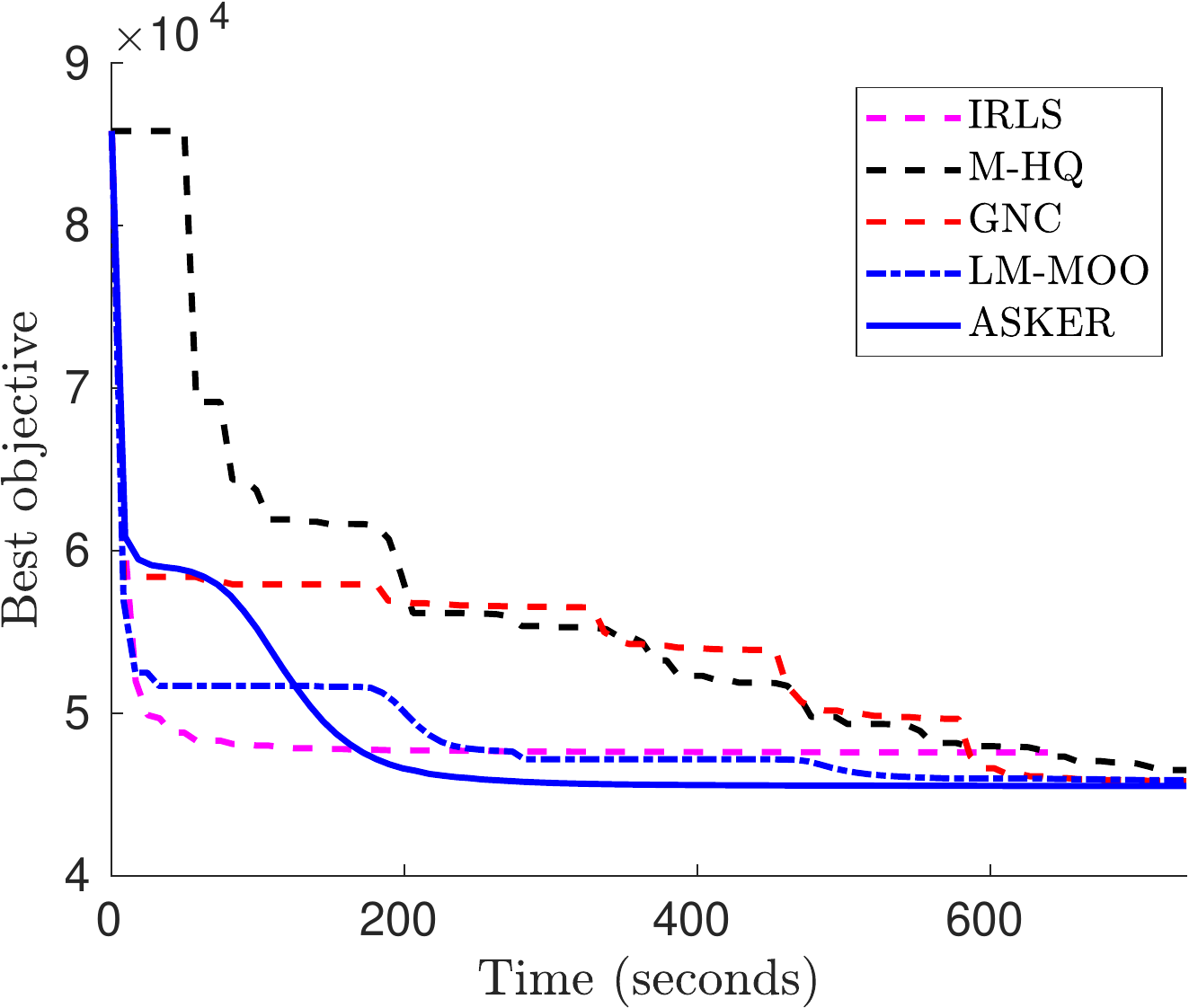}
        \caption{Final-394}
        \label{subfig:fin394}
    \end{subfigure}

    \caption{Performance of the algorithms. We compare ours (ASKER) against standard IRLS, M-HQ~\cite{zach2014robust}, GNC~\cite{zach2018descending}, and LM-MOO~\cite{zach2019pareto}, which are state-of-the-art methods. }
    \label{fig:results_convergence}
\end{figure*}

%% file: tex/table_inliers.tex
\begin{table}[ht]
\centering
\resizebox{1.0\columnwidth}{!}{
\begin{tabular}{|c|c|c|c|c|c|}
\hline
           &  IRLS    & M-HQ   & GNC   & MOO     & ASKER \\ \hline
Ladybug-49 &  80.4    &  \textbf{82.3}  & 82.1  & 81.9    & \textbf{82.3}  \\ \hline
Trafalgar-21 &  50.9    &  69.0  & 68.3  & 68.8    & \textbf{69.10}  \\ \hline
Trafalgar-201 &  66.31    &  69.2  & 69.1  & 68.8    & \textbf{69.33}  \\ \hline
Trafalgar-225 &  67.05    &  69.8  & 69.9  & \textbf{70.01}    & \textbf{70.01}  \\ \hline

\end{tabular}
}
\caption{Inlier percentage achieved by the methods.}
\label{table:results_inliers}
\end{table}

%% file: supp/more_results.tex
\section{Additional Experimental Results}
\input{supp/fig_new_results.tex}
In this section, we provide more experimental results for the robust bundle adjustment experiment. The list of 20 instances that are used for the experiments includes: Trafalgar-126, Trafalgar-138, Dubrovnik-150, Dubrovnik-16, Trafalgar-201, Dubrovnik-202, Trafalgar-21, Trafalgar-225, Dubrovnik-253, Ladybug-318,Final-93, Final-394, Venice-245, Dubrovnik-308, Dubrovnik-356, Venice-744, Venice-89, Venice-951, Trafalgar-39, Ladybug-49. Besides the results shown in the main manuscript, Figure~\ref{fig:results} plots the results for $8$ additional datasets.

%% file: supp/fig_new_results.tex
\begin{figure*}[htb]
    \centering
    \begin{subfigure}[b]{0.24\linewidth}
        \includegraphics[width=\textwidth]{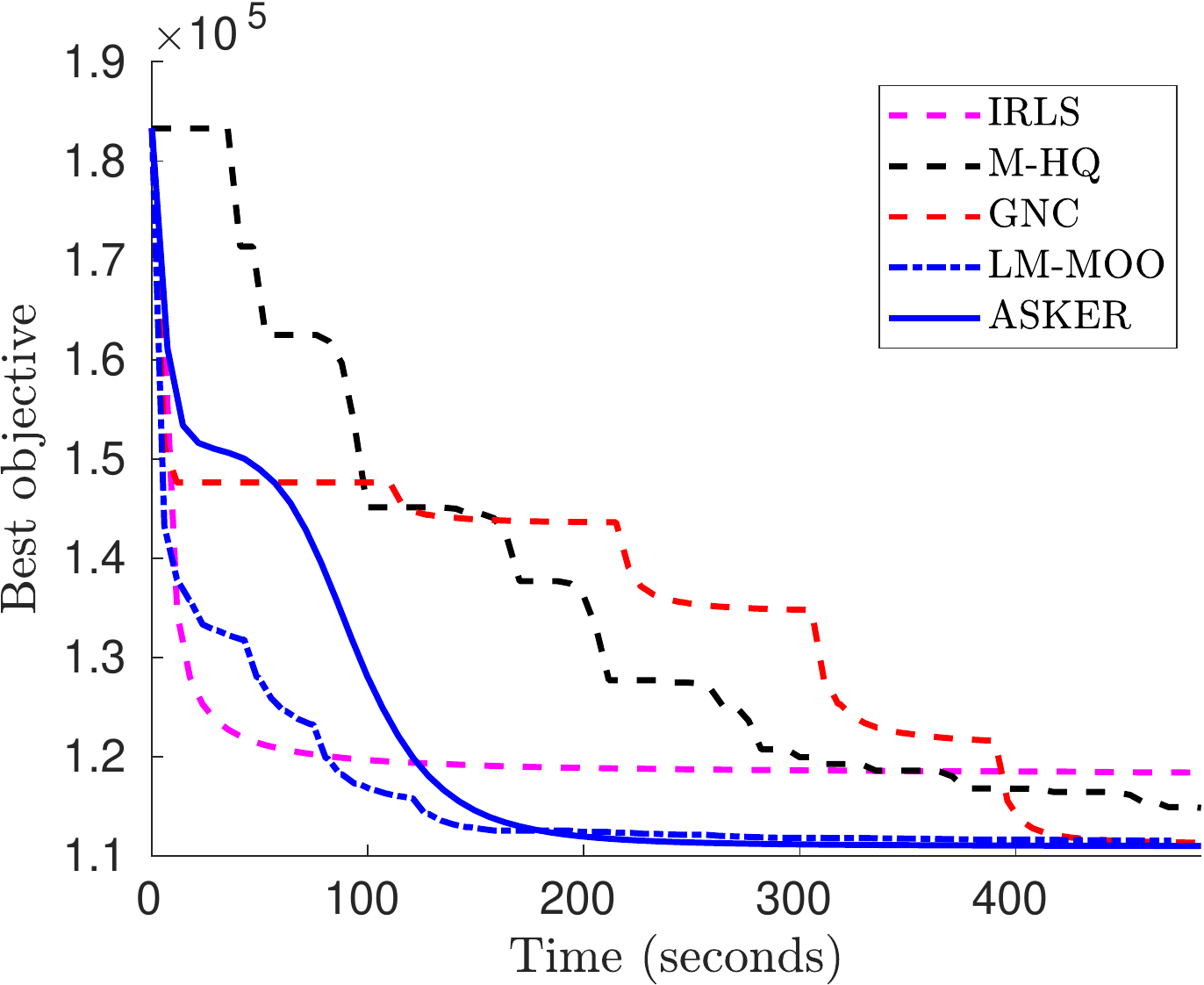}
        \caption{Venice-245}
        \label{subfig:tra126}
    \end{subfigure}
    \begin{subfigure}[b]{0.24\linewidth}
        \includegraphics[width=\textwidth]{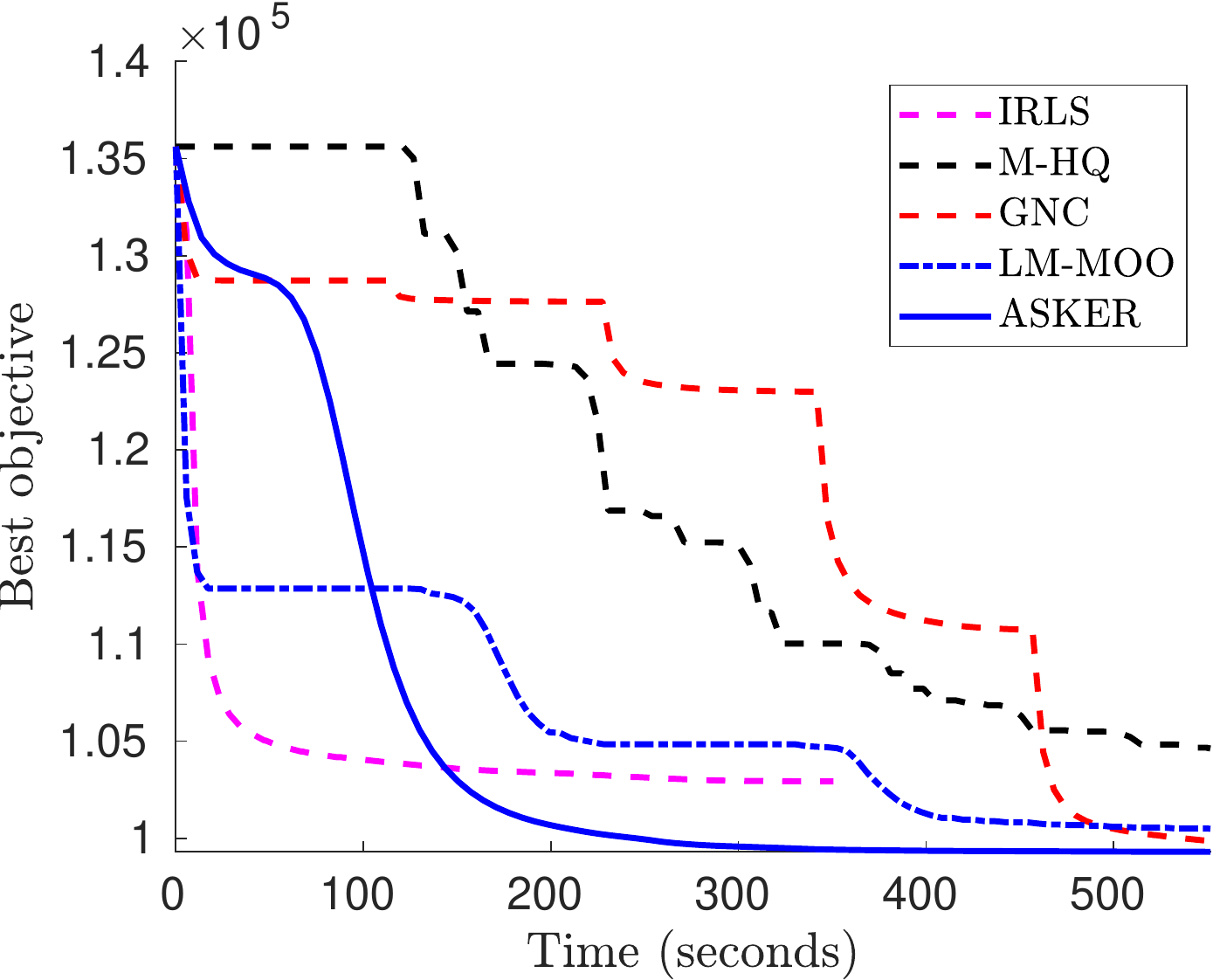}
        \caption{Dubrovnik-308}
        \label{subfig:tra126}
    \end{subfigure}
    \begin{subfigure}[b]{0.24\linewidth}
        \includegraphics[width=\textwidth]{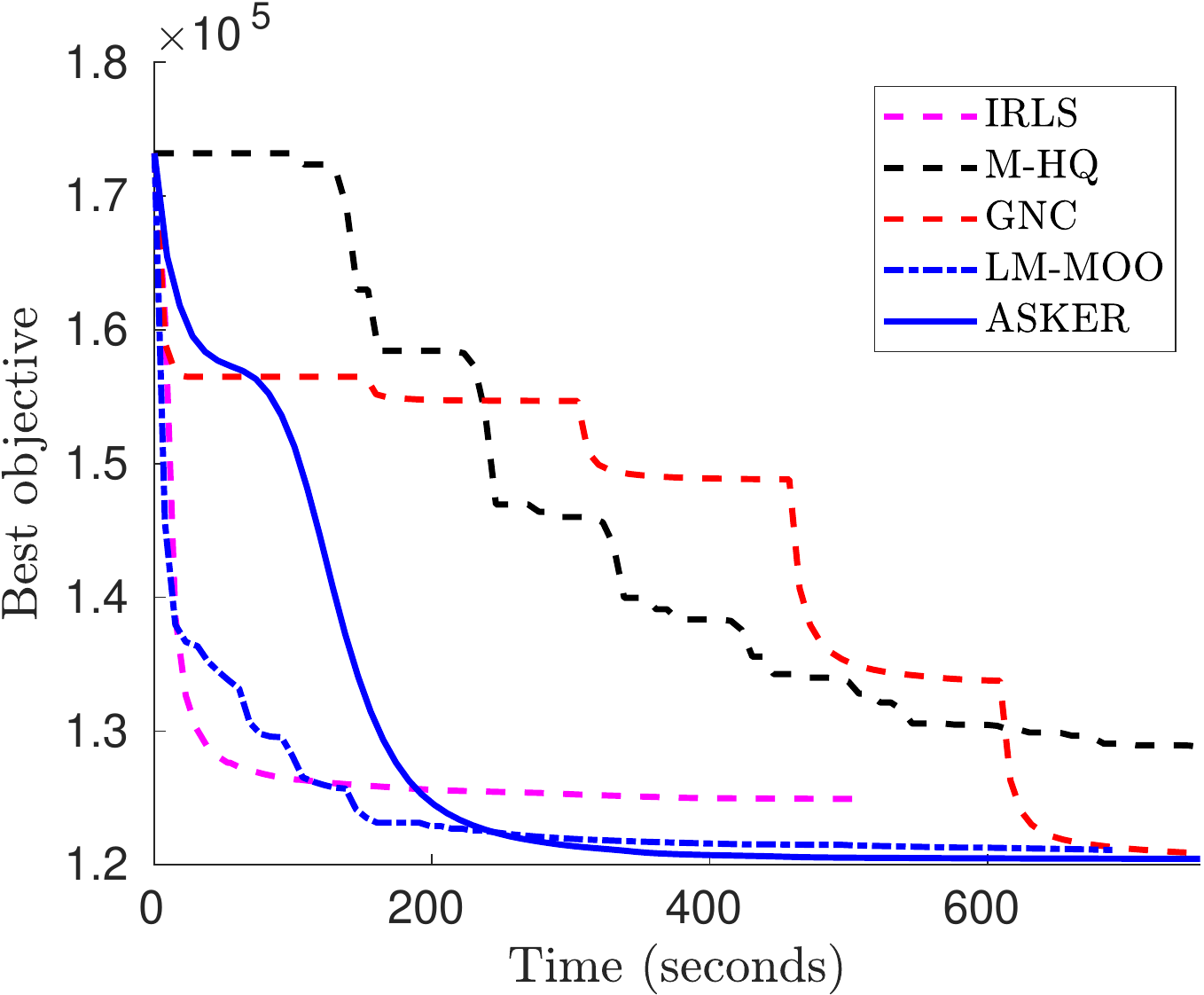}
        \caption{Dubrovnik-356}
        \label{subfig:tra126}
    \end{subfigure}
    \begin{subfigure}[b]{0.24\linewidth}
        \includegraphics[width=\textwidth]{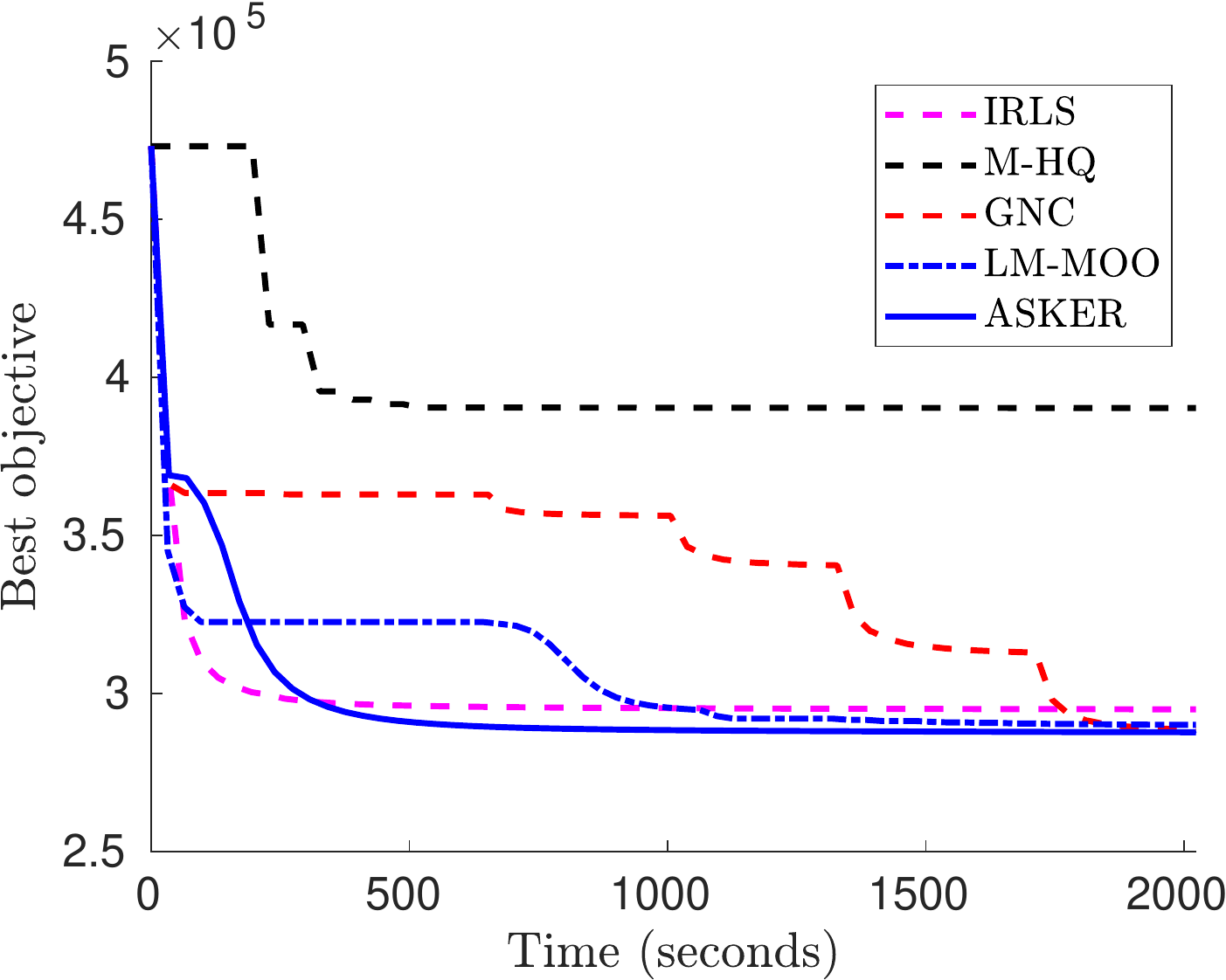}
        \caption{Venice-744}
        \label{subfig:tra126}
    \end{subfigure}
    \begin{subfigure}[b]{0.24\linewidth}
        \includegraphics[width=\textwidth]{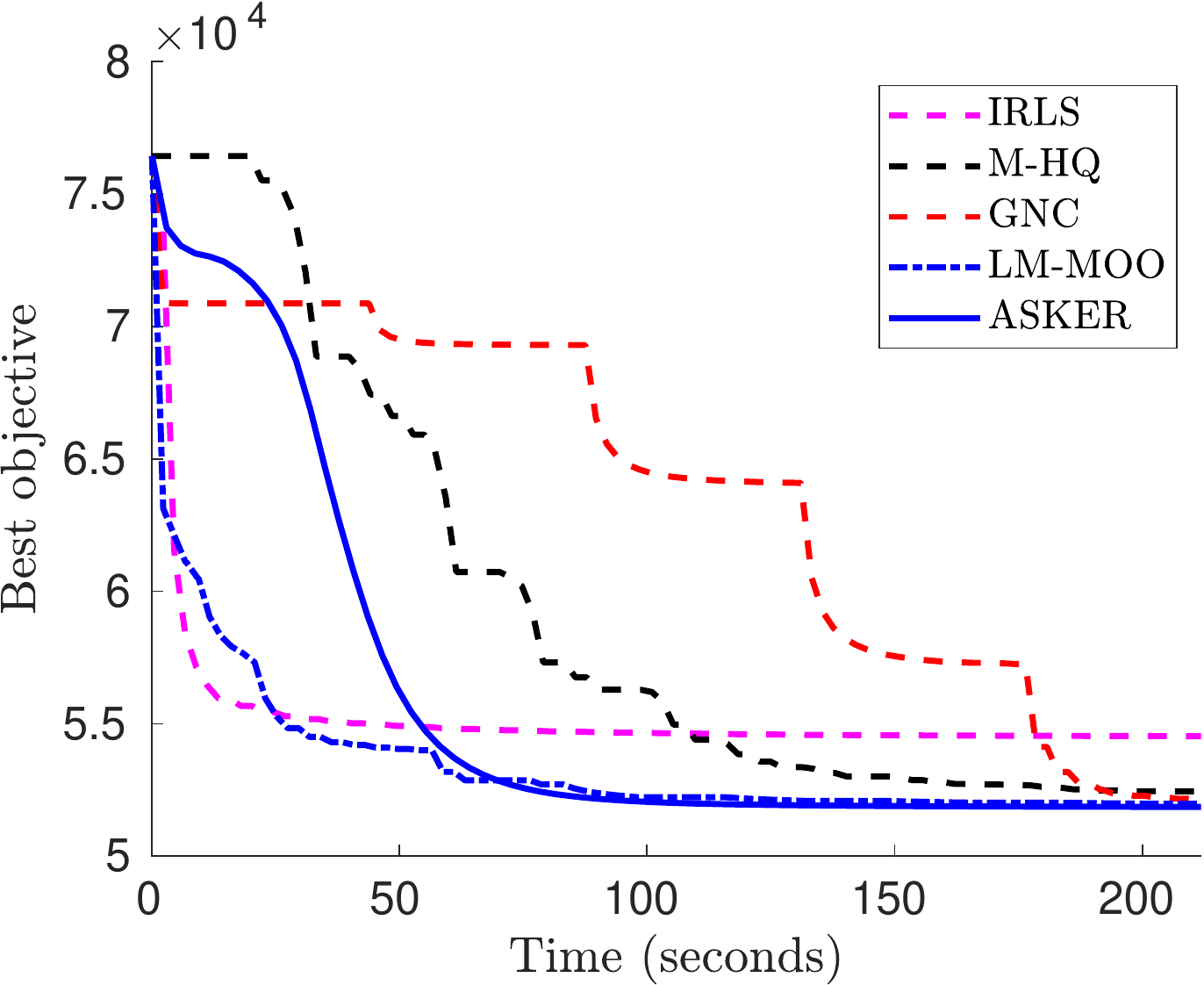}
        \caption{Venice-89}
        \label{subfig:tra126}
    \end{subfigure}
    \begin{subfigure}[b]{0.24\linewidth}
        \includegraphics[width=\textwidth]{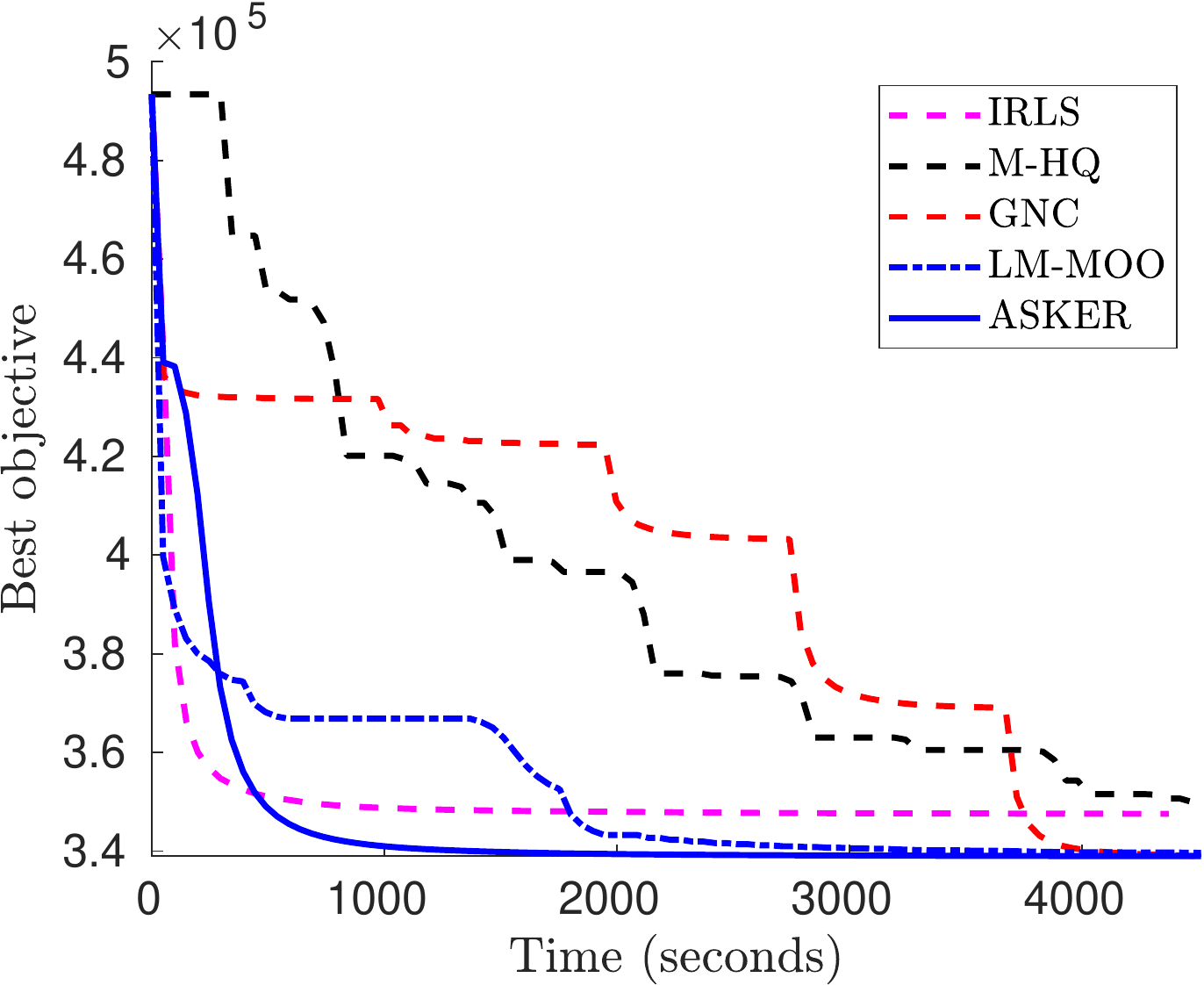}
        \caption{Venice-951}
        \label{subfig:tra126}
    \end{subfigure}
    \begin{subfigure}[b]{0.24\linewidth}
        \includegraphics[width=\textwidth]{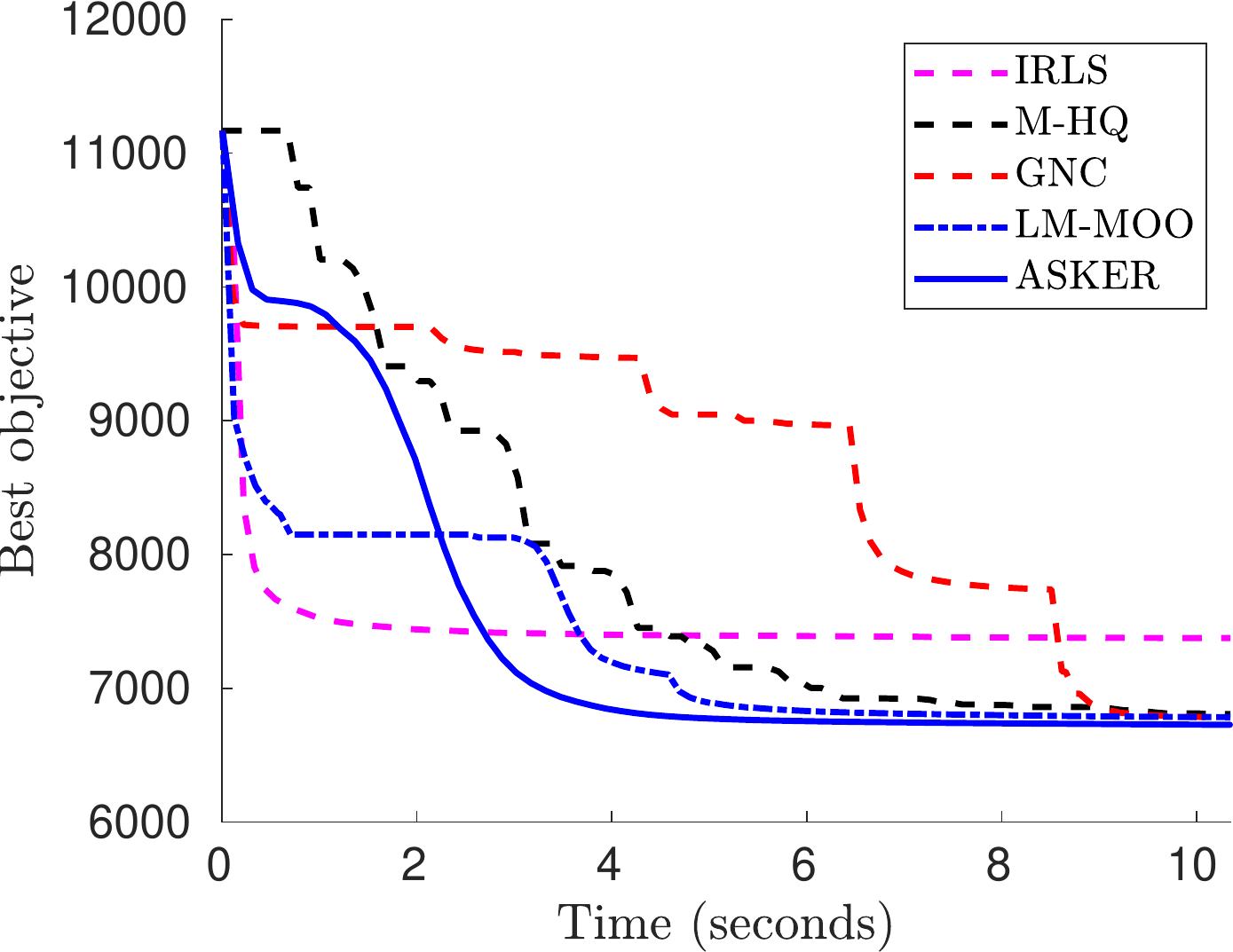}
        \caption{Trafalgar-39}
        \label{subfig:tra126}
    \end{subfigure}
    \begin{subfigure}[b]{0.24\linewidth}
        \includegraphics[width=\textwidth]{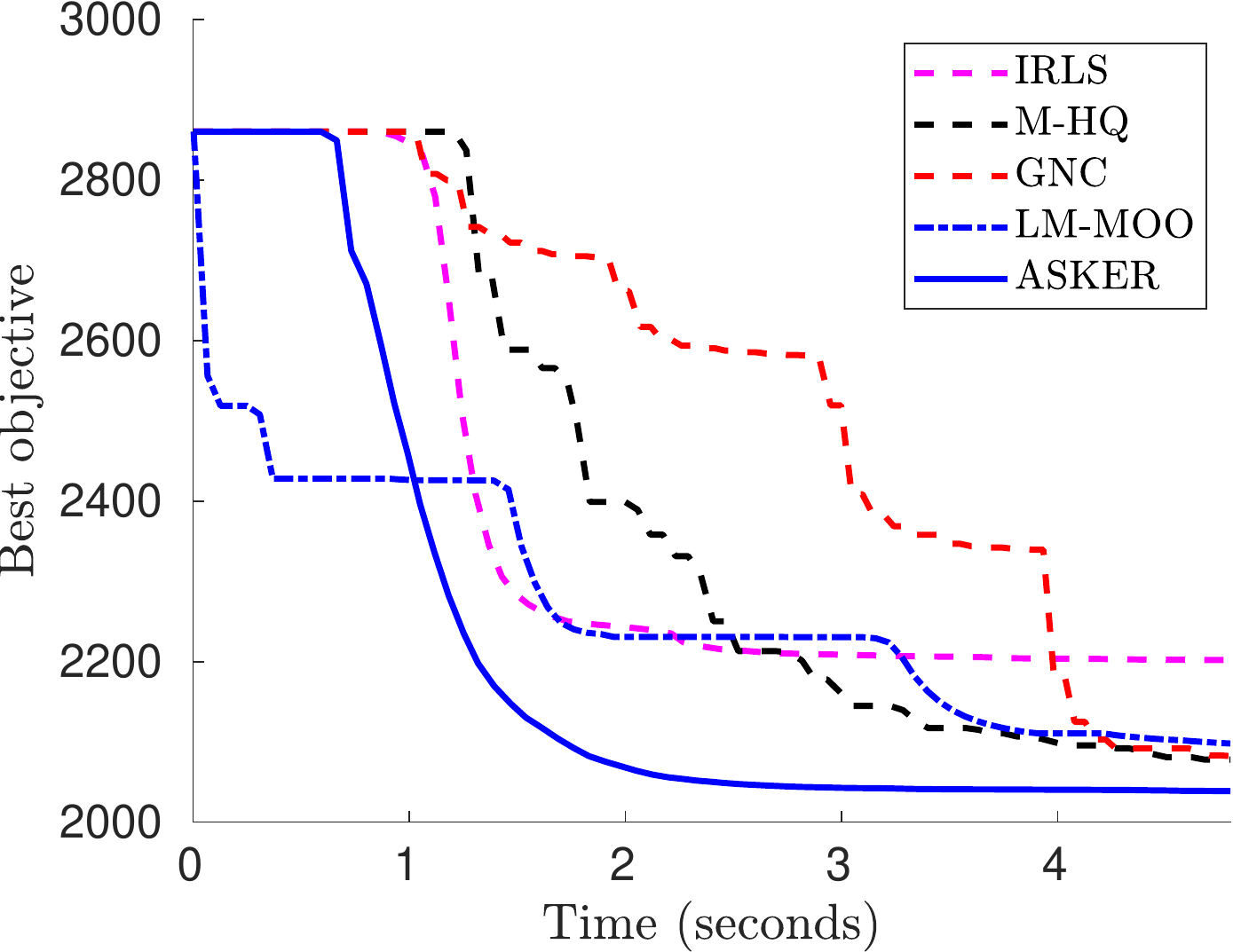}
        \caption{Ladybug-49}
        \label{subfig:tra126}
    \end{subfigure}
    \caption{Additional Results for Robust Bundle Adjustment}
    \label{fig:results}
\end{figure*}

%% file: supp/construction_visual.tex
\section{Reconstruction Results}
\begin{figure*}
    \centering
    \includegraphics[width = 0.45\textwidth]{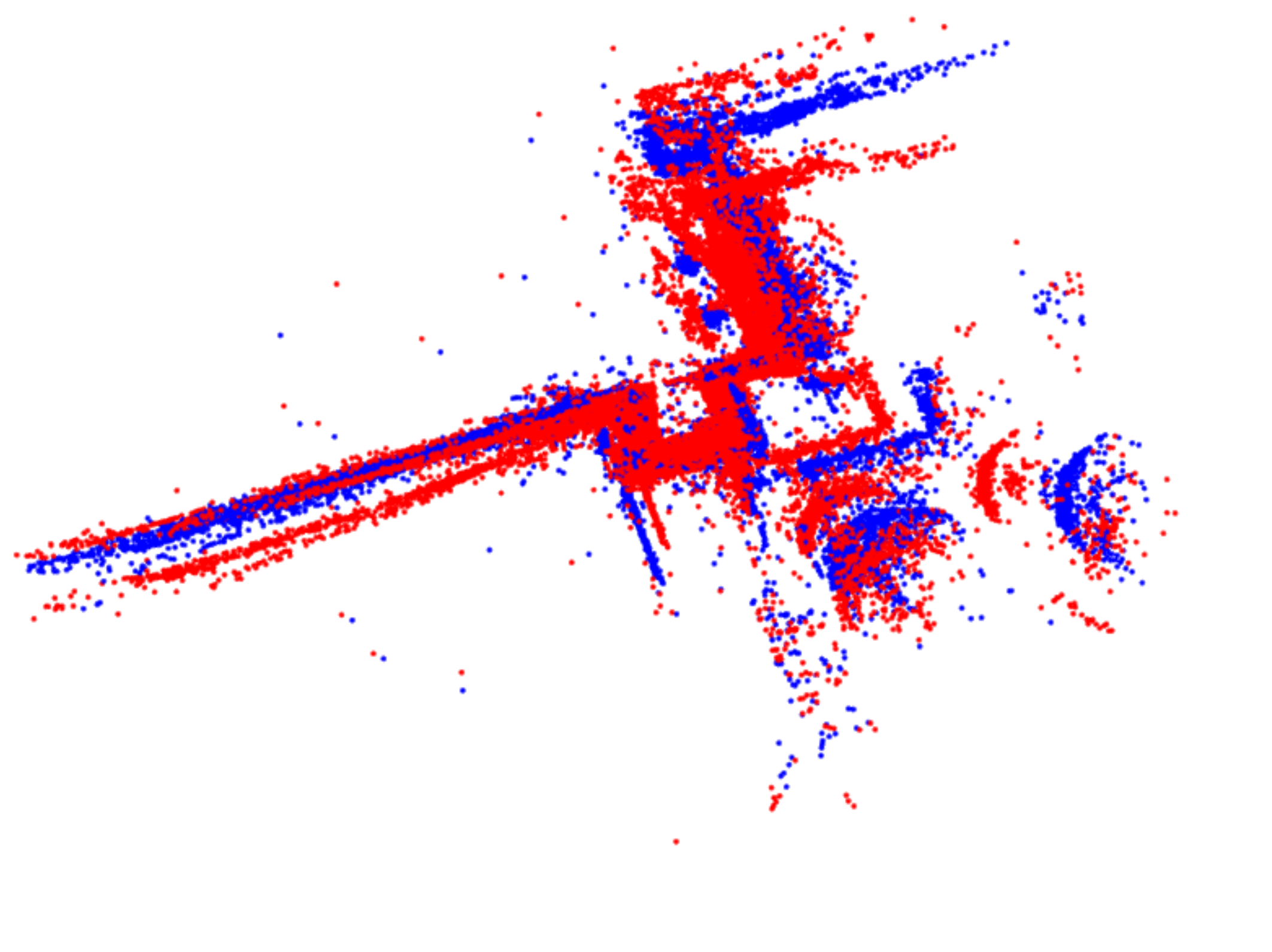}
    \includegraphics[width = 0.45\textwidth]{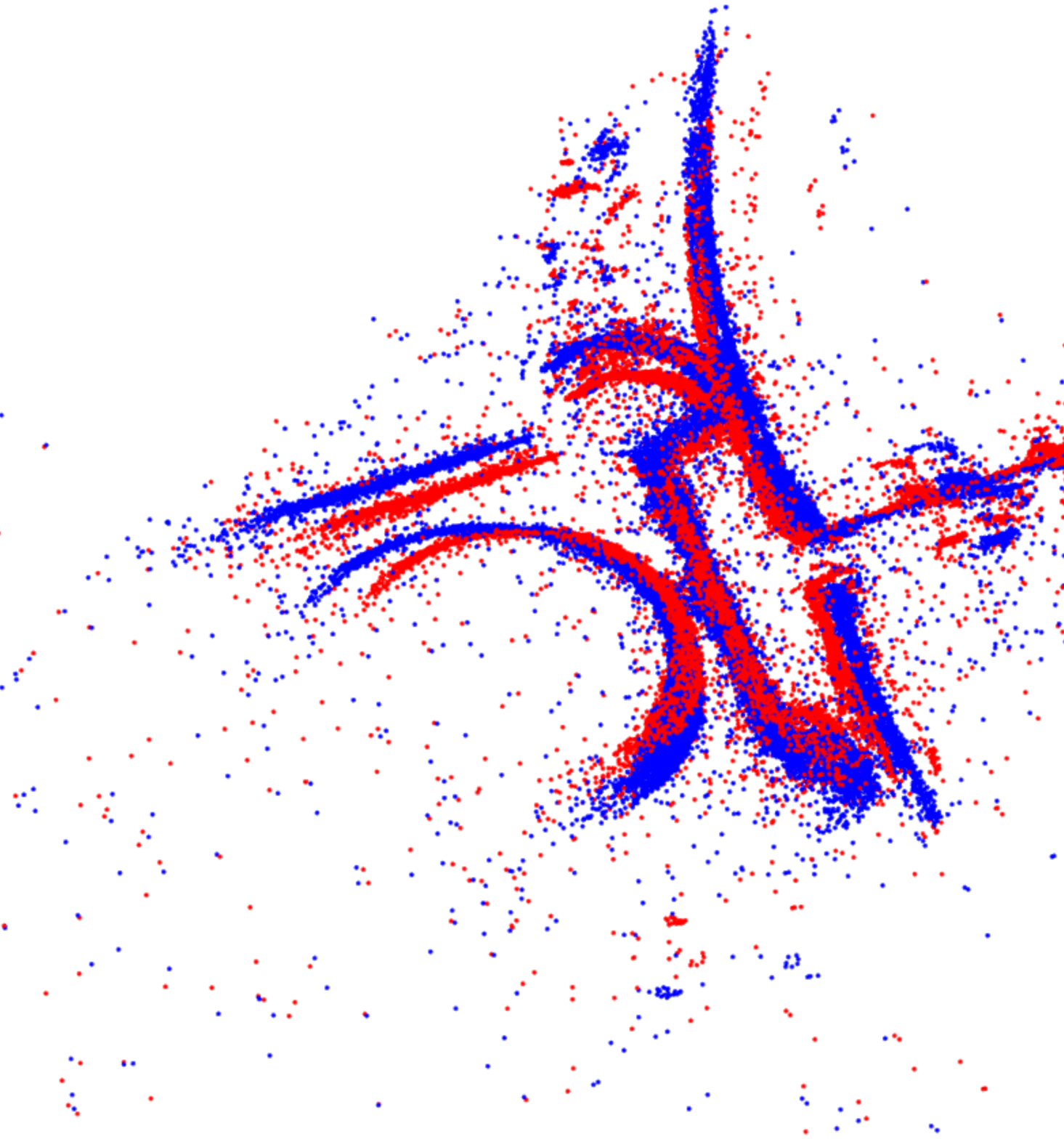}
    \caption{Top-down view of the reconstructed structures of Venice-89 (left) and Final-394 (right). Our results (\textcolor{blue}{ASKER}) are plotted in \textcolor{blue}{blue}, while \textcolor{red}{IRLS} results are shown in \textcolor{red}{red}. Observe that ASKER converges to better solutions, hence the 3D points form a better structure compared to IRLS.}
    \label{fig:reconstruction_results}
\end{figure*}

To demonstrate the concept of poor local minima, we show in this section the 3D reconstructed structures for two large datasets: Venice-89 and Final-394. The results are shown in Figure~\ref{fig:reconstruction_results}. Observe that by converging to lower robust costs, ASKER provides visually better structures compared to IRLS, which is easily trapped at a poor local minimum, resulting in higher objective values for most problem instances.

%% file: supp/step_computation.tex
\section{Detailed Algorithm for Step Computation}
In this section, we summarize the detailed algorithm for step computation (Line 10, Algorithm 1). Note that we choose the initial value of $\lambda$ to be $\lambda_{\text{init}} = 0.5$, and the values of $\lambda$ is modified according to Algorithm~\ref{alg:step_computation}. In particular, if $\bx^{t+1}$ is accepted to the filter, we reduce $\lambda$, allowing more exploration for the following steps. However, when $\bx^{t+1}$ returned by the cooperative step is not accepted, the value of $\lambda$ is reset to $\lambda_{\text{init}}$, and the restoration step is executed.
\input{supp/algo_step.tex}

%% file: supp/algo_step.tex
\begin{algorithm}[ht]\centering
\caption{Step Computation}
\label{alg:step_computation}                         
\begin{algorithmic}[ht]                   
	\REQUIRE Current value $\bx^t$, $\mu_f, \mu_g$, current damping value $\lambda^t$, $\lambda_{\text{init}} = 0.5$.
	\STATE Compute $\bg_f, \bg_h, \bH_f, \bH_h$ from $\bx^t$
	\STATE Compute $\Delta\bx$ using (13)
	\STATE $\bx^{t+1} \leftarrow \bx^{t} + \Delta\bx $
	\IF{$\bx^{t+1} \notin \bbF$} 
	    \STATE /*Perform Restoration Step*/
	    \STATE $\Delta\bx \leftarrow \gamma \begin{pmatrix} \mathbf{0} \\ -\bs\end{pmatrix}$, $\gamma$ is computed based on (15)
	    \STATE $\bx^{t+1} \leftarrow \bx^{t} + \Delta\bx $
	    \STATE $\lambda^{t+1} \leftarrow \lambda_{\text{init}}$
	\ELSE
	    \STATE $\lambda^{t+1}  \leftarrow \lambda^{t} / 10$
	\ENDIF
	
	\RETURN $\bx^t$
\end{algorithmic}
\end{algorithm}

%% file: supp/param_choice.tex
\section{Parameter Choices}
\begin{figure}[ht]
    \centering
    \includegraphics[width = 0.75\columnwidth]{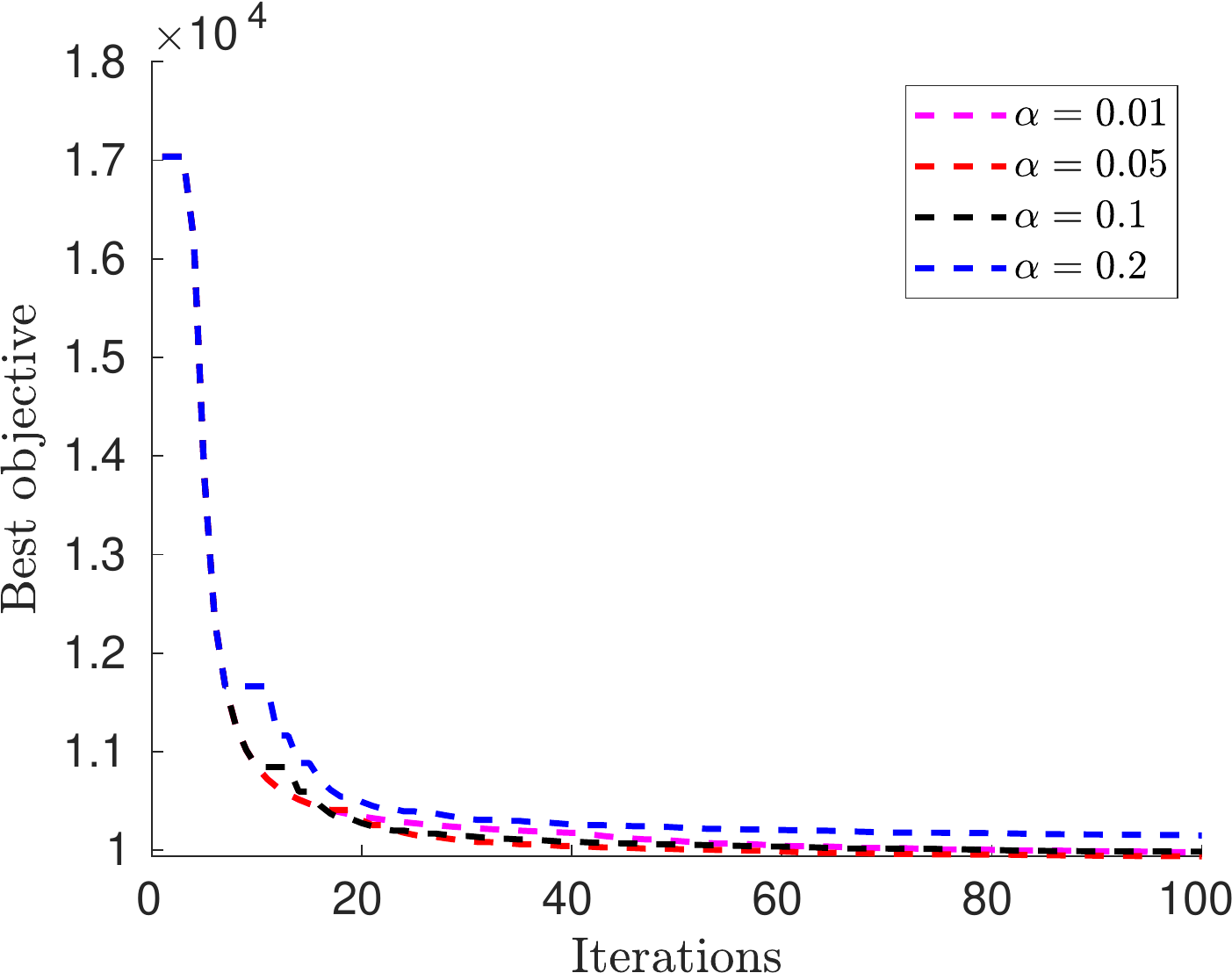}
    \caption{Convergence of algorithm for different values of alpha margin $\alpha$.}
    \label{fig:alpha_choice}
\end{figure}
This section provides some study on the effects of parameter choices to the performance of our algorithm. 
\subsection{Filter Margin $\alpha$ }
Figure~\ref{fig:alpha_choice} shows the evolution of the best cost with four different values of $\alpha$. Observe that our algorithm is insensitive to the choice of $0.01 \le \alpha \le 0.1$, as the converged solutions are similar. However, when $\alpha \ge 0.2$ the converged objectives is higher, since the margin becomes unreasonably large that prevents the meaningful reduction provided by the cooperative step. In most experiments, we found $\alpha = 10^{-4}$ provides the best results, but other values of can also be used depending on the application and the starting values of the constraint violation.
\subsection{Initial Scaling Values $s_i$}
\begin{figure}[ht]
    \centering
    \includegraphics[width = 0.85\columnwidth]{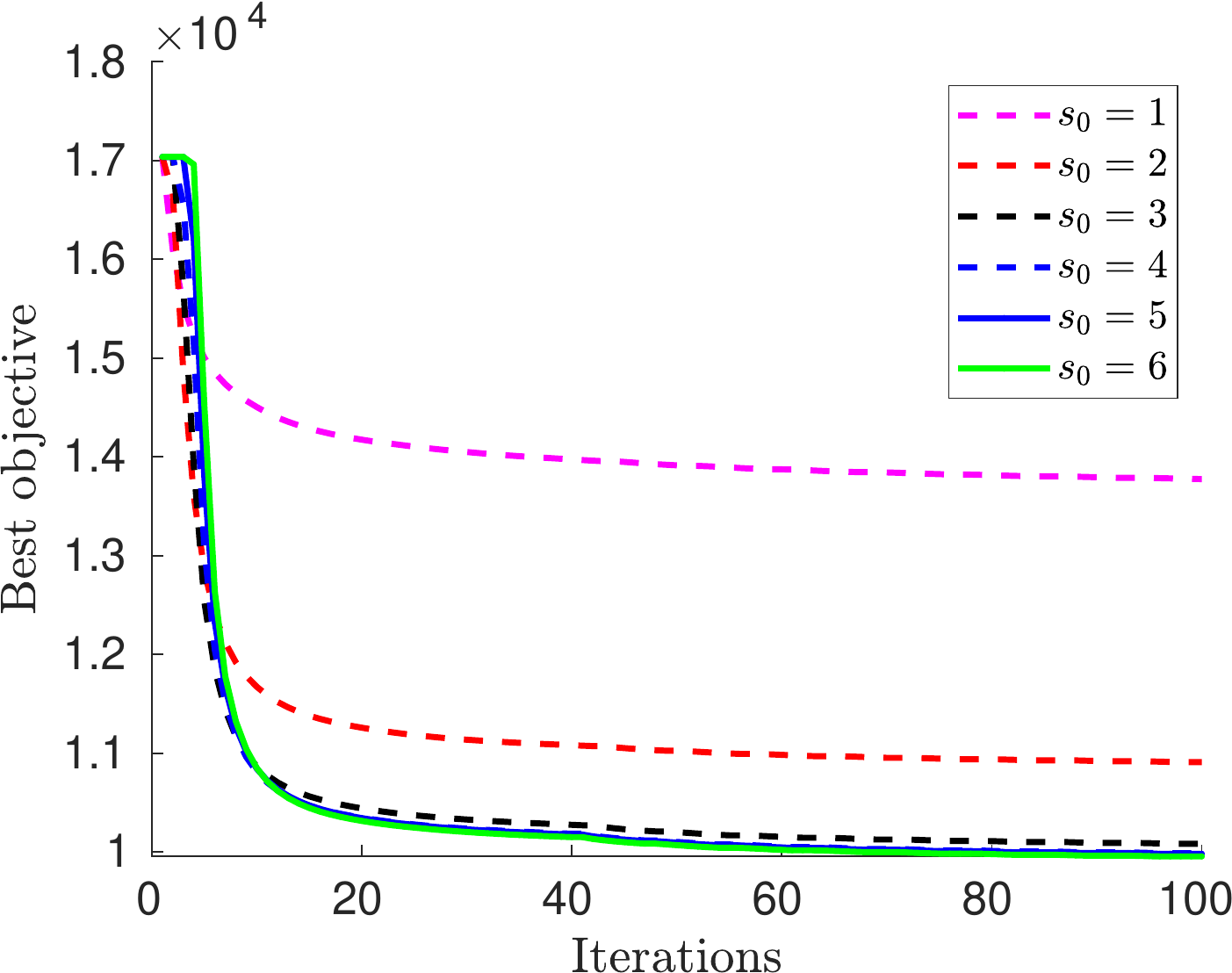}
    \caption{Convergence of algorithm for different initialization values of $s_i$.}
    \label{fig:s_init}
\end{figure}

In Figure~\ref{fig:s_init}, we show the performance of our algorithm under different initialization values for $s_i$. Observe that when $s_i$ are initialized with values $s_0$ that are less than $3$, the converged objectives are poor, because the kernels are not effectively scaled. When $s_i$ are initialized to $s_0 \ge 3$, the converged objectives are similar. Figure~\ref{fig:s_init} also demonstrates that our algorithm is also insensitive to the initialization. In particular, even though when $s_i$ are initialized to unnecessarily large values, the optimization process is still able to obtain competitive results within a few number of iterations.

%% file: supp/convergence.tex
\section{Convergence}
\begin{figure}[ht]
    \centering
    \includegraphics[width = 0.75\columnwidth]{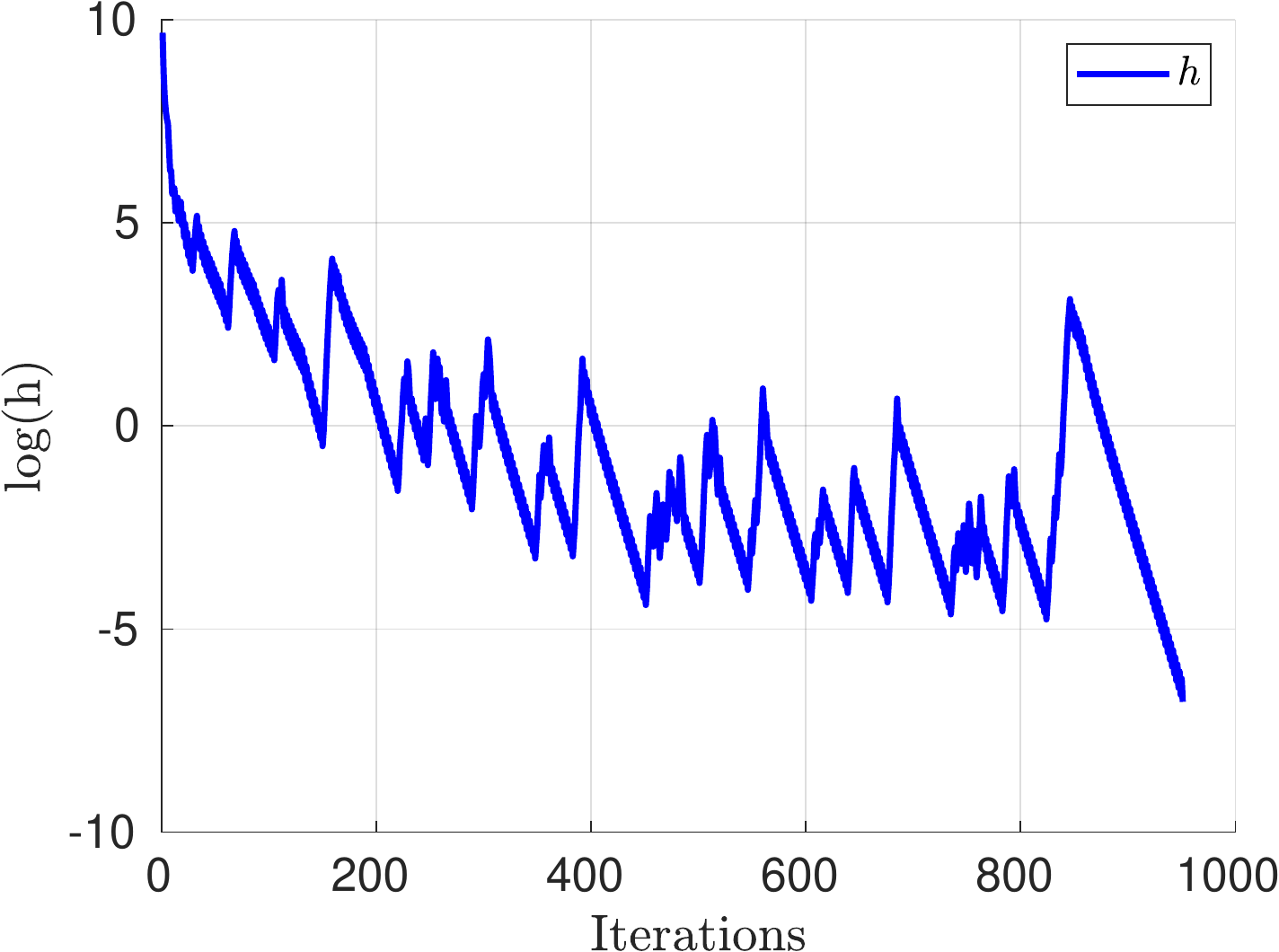}
    \includegraphics[width = 0.75\columnwidth]{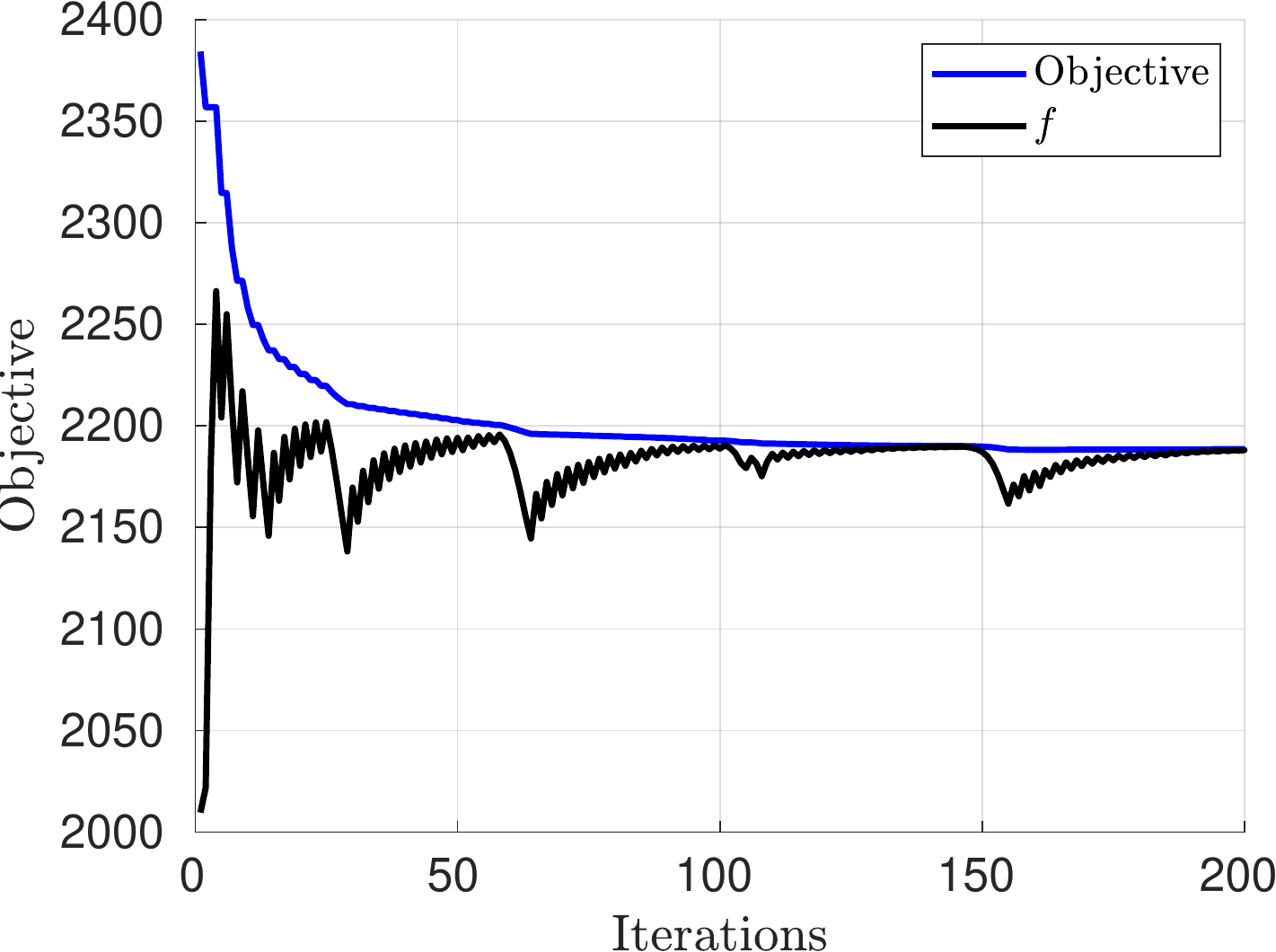}
    \caption{Convergence of algorithm to feasible region.}
    \label{fig:h}
\end{figure}
The convergence of our algorithm to $h=0$ and a stationary solution of $f$ is guaranteed by the nature of the filter algorithm~\cite{ribeiro2008global}. In Figure~\ref{fig:h}, we plots the objective of $h$ over the iterations (top), together with the evolution of the objective of the original function and $f$ (bottom). Observe that $f$ and $h$ do not monotonically increased or decreased, but vary during the optimization process, as long as the pairs $(f^t,h^t)$ are accepted to the filter. The objective of $h$ converges to $0$, while $f$ and the real objective converge to the same value. The exploration of $f$ and $h$ in the filter drives the original objective to a local optimal solution. Note that after a number of iterations, the $h$ becomes sufficiently small, and further exploration does not offer much improvement to the original objective. Therefore, one can define a early stopping criterion based on $h$. When all $s_i$ are set to $0$, our algorithm reverts to IRLS.

%% file: supp/gradient.tex
\section{Gradient Computation}

This section details the computation of $\bg_f$, $\bg_h$ and $\bH_f$, $\bH_h$ that are used in the step computation discussed in Section 5.2.1 of the main paper.

Recall the formulation of $f(\bx)$ and $h(\bx)$ 
\begin{align}
    f(\bx) = \sum_{i=1}^N \kernel \left(\frac{\|\br_i (\btheta)\|}{1 + s^2_i}\right) & &  h(\bx) = \sum_i s^2_i,
\end{align}
where $\bx = [\btheta^T \; \bs^T]^T$ and $\bs = [s_1 \dots s_N]^T$ as described in the main paper.

For brevity, let $\bhr_i$ denote the ``scaled" residual obtained by dividing the original residual by $1+s_i^2$, i.e.,
\begin{equation}
    \kernel(\|\bhr_i(\bx)\|) = \kernel \left(\frac{\|\br_i(\btheta)\|}{1+s_i^2}\right) = \kernel
    \left(\begin{Vmatrix}
                \frac{\br^1_i(\btheta)}{1+s_i^2} \\
                \vdots \\
                \frac{\br^p_i(\btheta)}{1+s_i^2}, 
            \end{Vmatrix} 
    \right)
\end{equation}
where $\br_i^j$ ($j=1 \dots p$) is the $j$-th element of the residual vector $\br_i \in \bbR^p$.

Also, consider the following first-order approximation,
\begin{equation}
    \kernel(\|\bhr_i(\bx + \Delta \bx)\|) =\kernel(\|\bhr_i(\bx) + \bJ_i\Delta \bx)\|),   
\end{equation}
where $\bJ_i$ is the Jacobian of $\bhr_i(\bx + \Delta \bx)$ w.r.t. $\Delta \bx$ evaluated at $\Delta \bx = \mathbf{0}$.

To employ non-linear least squares solvers for our method, at each step, we utilize a convex quadratic majorizer $\hat{\kernel}$ for each scaled residual vector $\bhr_i$ such that $\kernel(r) \le \hat{\kernel}(r)$. In this work, we make use of the well-known IRLS majorizer:
\begin{equation}
    \hat{\kernel}(\|\bx + \bJ_i\Delta\bx\|) = \frac{1}{2}\omega(\|\bhr_i\|)(\|\bhr_i + \bJ_i\Delta\bx\|^2 - \|\bhr_i\|^2) + \kernel(\|\bhr_i\|)
    \label{eq:majorizer}
\end{equation}
where $\bhr_i$ is a short hand notation for $\bhr_i(\bx)$, and $\omega(r)$ with $r \in \bbR_+$ is defined as
\begin{equation}
    \omega(r) := \frac{\kernel'(r)}{r},
\end{equation}
which acts as the weight for the residuals. 

For brevity, let $\omega_i$ denote $\omega(\|\bhr_i\|)$. The equation~\eqref{eq:majorizer} can be rewritten as 
\begin{equation}
    \hat{\kernel}(\|\bx + \bJ_i\Delta\bx\|) = \frac{1}{2}\|\sqrt{\omega_i}\bhr_i + \sqrt{\omega_i}\bJ_i\Delta\bx\|^2 - \frac{\omega_i}{2}\|\bhr_i\|^2 + \kernel(\|\bhr_i\|)
    \label{eq:majorizer_1}
\end{equation}
Based on~\eqref{eq:majorizer_1}, we can define a new Jacobian matrix and residual function for the function $\hat{\kernel}$ 
\begin{align}
    \tilde{\br}_i = \sqrt{\omega_i}\bhr_i && \tilde{\bJ}_i = \sqrt{\omega_i} \bJ_i,
\end{align}
Then, under the LM framework, the gradient $\bg_f$ and the approximated Hessian $\bH_f$ can be computed as
\begin{align}
    \bg_f = \sum_i \tilde{\bJ}_i^T\tilde{\br}_i & & \bH_f = \sum_i \tilde{\bJ}_i^T\tilde{\bJ}_i
\end{align}
For the case of $h$, its gradient and Hessian can be computed in closed form:
\begin{align}
    \bg_h = 2[\mathbf{0} \;\; \bs ]^T & & \bH_h =
    2\begin{bmatrix}
    \begin{aligned}
    &\mathbf{0} && \mathbf{0} \\
    & \mathbf{0} && \mathbf{I}_{N\times N} 
    \end{aligned}
    \end{bmatrix}
\end{align}
In our experiments, to obtain better results, in the cooperative step, we increase the damping for $\bH$ to decrease the convergence rate of $h(\bx)$ (to zero). In particular, we use 
\begin{equation}
    \bH_h =
    2\begin{bmatrix}
    \begin{aligned}
    &\mathbf{0} && \mathbf{0} \\
    & \mathbf{0} && (1+\lambda_h)\mathbf{I}_{N\times N} 
    \end{aligned}
    \end{bmatrix}
\end{equation}
where $\lambda_h$ is initially set to $2$. During the iterations, $\lambda_h$ is decreased by $\lambda_h \leftarrow 0.9\lambda_h$ if the new solution is accepted by the filter, otherwise reset to  $\lambda_h \leftarrow 2$.